\pdfoutput=1

\documentclass[11pt]{article}

\usepackage[final]{acl}

\usepackage{times}
\usepackage{latexsym}
\usepackage[T1]{fontenc}
\usepackage[utf8]{inputenc}
\usepackage{microtype}
\usepackage{enumitem}
\usepackage{inconsolata}
\usepackage{graphicx}
\usepackage{caption} 
\usepackage{algorithm}
\usepackage{algorithmic}
\usepackage[fleqn]{amsmath}
\usepackage{xspace}
\usepackage{threeparttablex}
\usepackage{adjustbox}
\usepackage{multirow}
\usepackage{ amssymb }
\usepackage{newfloat}
\usepackage{listings}
\usepackage{tikz}
\usepackage{pifont}
\usepackage{booktabs}
\usepackage{xcolor}
\usepackage{array}
\usepackage{hyperref}     
\usepackage{url} 
\usepackage{amsfonts}  
\usepackage{nicefrac} 
\usepackage{array}
\usepackage{float}
\usepackage{subcaption}

\newcommand{\PreserveBackslash}[1]{\let\temp=\\#1\let\\=\temp}
\newcolumntype{C}[1]{>{\PreserveBackslash\centering}p{#1}}
\newcolumntype{R}[1]{>{\PreserveBackslash\raggedleft}p{#1}}
\newcolumntype{L}[1]{>{\PreserveBackslash\raggedright}p{#1}}
\definecolor{darkGreen}{HTML}{008000}
\definecolor{darkBlue}{HTML}{00008B}
\definecolor{darkRed}{HTML}{8B0000}

\usepackage{color}
\newcount\Comments  
\newcommand{\kibitz}[2]{\ifnum\Comments=1{\color{#1}{#2}}\fi}

\newcommand{\gnn}{\textsc{Gnn}\xspace}
\newcommand{\gnns}{\textsc{Gnn}s\xspace}

\newcommand{\graph}{\mathcal{G}}
\newcommand{\nodes}{\mathcal{V}}
\newcommand{\edges}{\mathcal{E}}
\newcommand{\ntypes}{\mathcal{A}}
\newcommand{\etypes}{\mathcal{R}}

\newcommand{\realset}{\mathbb{R}}

\newcommand{\OUT}[1]{}

\newcommand{\model}{\textsc{GraphTrex}\xspace}
\newcommand{\spanmodel}{\textsc{SpanTrex}\xspace}
\newcommand{\trex}{\textsc{Trex}\xspace}

\title{Temporal Relation Extraction in Clinical Texts: A Span-based Graph Transformer Approach}
\author{
  Rochana Chaturvedi \quad Peyman Baghershahi \quad Sourav Medya \quad Barbara Di Eugenio \\
  University of Illinois Chicago \\
  Chicago, IL, U.S.A. \\
  \texttt{\{rchatu2, pbaghe2, medya, bdieugen\}@uic.edu} \\
}

\begin{document}
\maketitle
\begin{abstract}
Temporal information extraction from unstructured text is essential for contextualizing events and deriving actionable insights, particularly in the medical domain. We address the task of extracting clinical events and their temporal relations using the well-studied I2B2 2012 Temporal Relations Challenge corpus. This task is inherently challenging due to complex clinical language, long documents, and sparse annotations. We introduce \model, a novel method integrating span-based entity-relation extraction, clinical large pre-trained language models (LPLMs), and Heterogeneous Graph Transformers (HGT) to capture local and global dependencies. Our HGT component facilitates information propagation across the document through innovative global landmarks that bridge distant entities and improves the state-of-the-art with \textit{5.5\%} improvement in the tempeval $F_1$ score over the previous best and up to \textit{8.9\%} improvement on long-range relations, which presents a formidable challenge. We further demonstrate generalizability by establishing a strong baseline on the E3C corpus. Not only does this work advance temporal information extraction, but also lays the groundwork for improved diagnostic and prognostic models through enhanced temporal reasoning.
\end{abstract}

\section{Introduction}
Electronic Health Records (EHRs) contain a wealth of information in both structured data and unstructured free-text notes written by healthcare providers. These notes are the preferred information source of physicians as they contain valuable insights often missing from the structured data \citep{hersh2013caveats, capurro2014availability}, including temporal progression of symptoms, diagnoses, and treatments. Modeling such temporal relations is important for understanding the underlying disease pathology. For example, a drug given \textit{after} a symptom appears may indicate a treatment, while a symptom appearing \textit{after} a drug may indicate an adverse effect. 
Precise extraction of event chronology from text can aid precision medicine by improving temporal reasoning for downstream applications \citep{chen2dataset, vashishtha2020temporal}.

One such application is facilitating \textit{opportunistic screening} of chronic health conditions such as Diabetes Type 2 (T2D) \cite{pickhardt2021opportunistic,scheetz2021real}.
Many underprivileged patients forgo regular check-ups but interact with the healthcare system for other reasons \cite{zhang2015novel,dhippayom2013}. Opportunistic screening\textemdash screening patients for certain conditions whenever they come in contact with the healthcare system, possibly for unrelated reasons\textemdash is expensive at scale. Leveraging patient histories from the clinical notes can enable this by identifying early risk factors, thereby enhancing healthcare equity and reducing clinicians' manual review burden. This requires accurate temporal relation extraction (\trex) to determine when risk factors emerge in a patient’s history. Beyond risk factors, \trex also plays a crucial role in understanding patterns of drug use and misuse, as well as assessing the long-term benefits and potential side effects of medications. For instance, the  shared task by \citet{yao-etal-2024-overview} on extracting cancer treatment trajectories highlights the growing significance of this vital research area. Modeling events and their temporal connections has shown promise for event forecasting in other fields \cite{li2021future}, but no comparable work has been applied to the clinical domain. While some patient representation models consider coarse temporal chronology across patient visits \citep{ghassemi2015multivariate, lee2020harmonized, chaturvedi2023sequential}, fine-grained temporal information in the clinical notes remains underutilized \cite{tvardik2018accuracy}.


\begin{figure*}[htbp]
\centering
\includegraphics[width=.9\linewidth]{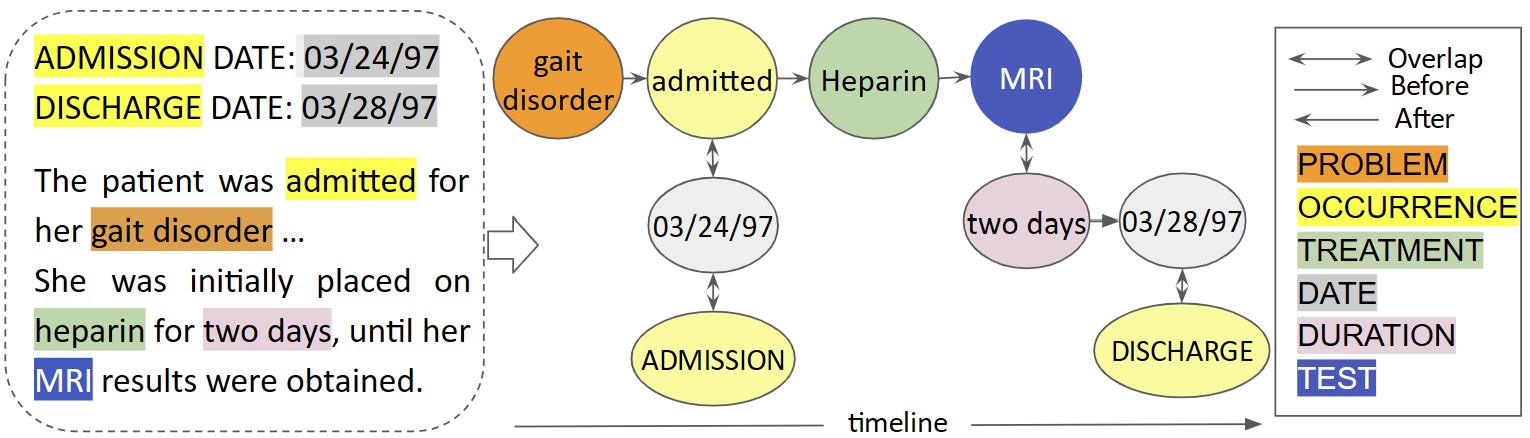} 
\caption{Temporal graph of clinical events and time expressions from a clinical note with color-coded spans as nodes and edges indicating their temporal relations.} \label{fig:temporal_kg}
\vspace{-2.5mm}
\end{figure*}

The clinical \trex task faces at least three challenges. First, clinical text requires domain-specific models due to its specialized vocabulary (e.g., `b.i.d.' meaning twice a day for dosage), unique temporal expressions (e.g., `three days postop'), and ambiguity in abbreviations across providers (e.g. `DM2' can mean Diabetes Mellitus Type 2 or Myotonic Dystrophy Type 2), requiring contextual interpretation. However, publicly available, moderate-sized, free or affordable corpora of clinical notes annotated with temporal relations are scarce due to costly expert annotation and strict privacy regulations. Several works use the I2B2 2012 corpus \cite{sun2013annotating, sun2013evaluating}, as we do here. Another corpus, THYME \cite{bethard-etal-2017-semeval, styler2014temporal},  requires a membership fee similar to the Linguistic Data Consortium (LDC). Fortunately, a small corpus annotated with THYME guidelines, E3C \cite{magnini2022european}, was recently released, and we report results on it as well. 

Second, \trex 
is often tackled using a sequential (pipeline) approach that first extracts entities and then classifies the entity pairs into a relation type \citep{xu2013end}. Unlike other relation extraction tasks, temporal relations are constrained by temporal logic and often implied through linguistic cues (verb tenses, temporal prepositions) and reasoning over the extracted relation pairs. Conventional sequential approaches struggle to capture such interactions between entities and temporal relations, often leading to error propagation. 

Third, clinical notes generally comprise long documents with extensive patient history, findings, treatment plans, etc. However, the recent transformer-based models that perform exceptionally well in NLP tasks have context-length limitations. Due to this, recent works in clinical \trex target entity pairs across small context lengths, such as only event pairs across three consecutive sentences \citep{han2020domain} to avoid the quadratic complexity of document-level context, overlooking both the long-distance pairs and interactions with the time expressions. Consider the example in Figure~\ref{fig:temporal_kg}, containing a small excerpt of a real clinical note and corresponding temporal graph. It is easy to infer that gait disorder $<$ \textit{(before)} admission $<$ heparin $<$ MRI, by observing the local context. Given these patterns, one can also infer the long-range relation that gait disorder $<$ MRI. Inferring such long-range relations is essential to obtain more accurate timelines; however, it is known to be a formidable challenge \cite{qin-etal-2023-nlp}.

\paragraph{Contributions.} 
To address these challenges, we introduce  \model (\textbf{Graph}-based \textbf{T}emporal \textbf{R}elation \textbf{Ex}traction), an end-to-end framework integrating text- and graph-based methods, enabling a more structured and context-aware extraction. \model constructs a document-level Heterogeneous Graph (HG) from span-based model predictions using a clinical Pre-trained Language Model (PLM), then applies Heterogeneous Graph Transformers (HGT) for local and global information propagation. By leveraging tailored node types and heterogeneous information aggregation, \model achieves state-of-the-art performance on the I2B2 2012 corpus and establishes a competitive novel baseline on the E3C corpus. Importantly, our ablation experiments demonstrate that \model excels in inferring long-distance relations, a critical challenge in clinical \trex.

\section{Related Work}
Early clinical \trex approaches use a hybrid of rule-based and machine-learning-based systems that require extensive feature engineering and additional annotations \citep{xu2013end, tang2013hybrid, sohn2013comprehensive}. Transformer-based PLMs such as BERT \citep{bert2019} improve both entity extraction \citep{alsentzer2019publicly,si2019enhancing} and relation classification \citep{zhou2021clinical} to classify known gold-standard entity pairs, excluding unrelated pairs. However, they have not been applied to the full end-to-end extraction task due to document-level complexity, and remain limited to nearby event-event pairs \cite{han2020domain}. Since this task remains essential for real clinical use cases, we refocus on the complete document-level, end-to-end \trex using sliding windows and graph-based modeling to incorporate long contexts.

Contemporary clinical \trex methods have not shown improvement over the older baselines. Most of the works adopt traditional token-based models for entity extraction \citep{liu2017entity}, which are prone to cascading label misclassifications \citep{yu2022s}. Secondly, the pipeline approaches for end-to-end relation extraction struggle to capture the complex interactions between entities and their relationships. In the general domain, joint span-based approaches \citep{dixit2019span, lai2021joint, eberts2020span, yan2022empirical} have been shown to address these limitations by modeling entity and relation extraction together. They also mitigate the cascading error issue of token-based systems by enumerating all possible contiguous spans and then directly classifying them. Another joint framework REBEL \citep{huguet-cabot-navigli-2021-rebel-relation} that generates entity-relation triplets autoregressively has shown strong performance in document-level relation extraction. However, it underperforms in clinical \trex \citep{saiz2023end}, likely due to the unique challenges posed by temporal dependencies. We integrate a span-based approach with structured graph reasoning to address these challenges.\footnote{See Appendix \ref{app:additional_literature} for additional related works. }
\section{Problem Statement \& Data}
\subsection{Problem Statement}
Given a clinical note, we extract a temporal graph $\graph$, jointly identifying the nodes and edges. The nodes correspond to entities, which comprise either clinical events such as symptoms or medications; or time expressions (timex) such as dates, frequencies, or durations. Edges represent temporal relations. As shown in Figure \ref{fig:temporal_kg}, a directed edge from $e_1$ to $e_2$ represents the temporal relation $e_1$ \textit{Before} ($<$) $e_2$ or $e_2$ \textit{After} ($>$) $e_1$ while an undirected edge represents $e_1$ \textit{Overlaps} ($=$) $e_2$.  

\OUT{The end-to-end entity-relation task can be split into entity extraction (NER) and relation extraction (RE) steps. Entity extraction can be viewed as entity identification or span identification (EI) followed by entity classification (EC). Some works combine both by predicting a `none' type besides the available entity classes. The `none' type identifies text spans that are not identified as a participating entity. The relation extraction (RE) task can similarly be split into relation identification (RI) to identify entity pairs that participate in a relation and relation classification (RC) which identifies the relation type between the identified pairs. }


\subsection{Data}
We primarily focus on the I2B2 2012 Temporal Relations Challenge corpus \citep{sun2013annotating, sun2013evaluating} in this study, which features a substantial number of cross-sentence annotations, unlike other clinical \trex corpora. 
The data comprises 310 discharge summaries in English where entities include: 
\\\textbf{Clinical Events}: Events include six types\textemdash, namely TREATMENT, TEST, PROBLEM, CLINICAL DEPARTMENT, EVIDENTIAL, and OCCURRENCE. While TREATMENT, TEST, and PROBLEM are self-explanatory; EVIDENTIAL denotes information source (e.g. `tested', `complained'); OCCURRENCE refers to activities such as transfers between departments, admission, discharge, or follow-up. CLINICAL DEPARTMENT anchors key services in the patient's timeline, often reflecting changes in severity (e.g., ICU admission indicates higher severity).
\\\textbf{Time Expression (TimEx)}: Includes DATE, TIME, DURATION, and FREQUENCY. 
\\\textbf{SECTIME}: Time anchors marking section creation times, like ADMISSION (clinical history) or DISCHARGE (hospital course) dates.
\\ \textbf{Temporal relations or Tlinks:} Include \textit{Before} ($<$), \textit{After} ($>$), and \textit{Overlap} ($=$). TLinks can exist between event-event (EE), event-timex or event-sectime (ET), and timex-timex (TT).
\\
\textit{Train-test Split.}
The data is divided into 190 training files and 120 test files. Following \citep{zhou2021clinical}, we sample 9 files from the training set as the development set. On average, a document contains 86.6 events, 12.4 timexes, and 176 TLinks. There's an average of 576 tokens per document, 110 entities, and 197 annotated relations.

We also conduct initial experiments on the English subset of the E3C corpus \citep{magnini2022european}, a collection of clinical cases. The dataset includes 48 test documents and 36 training documents, with 7 (20\%) reserved for development. 

Appendix \ref{app:data} contains additional data details.

\section{Our Method: \model}
\begin{figure*}[ht]
    \centering
    \includegraphics[width=\linewidth]{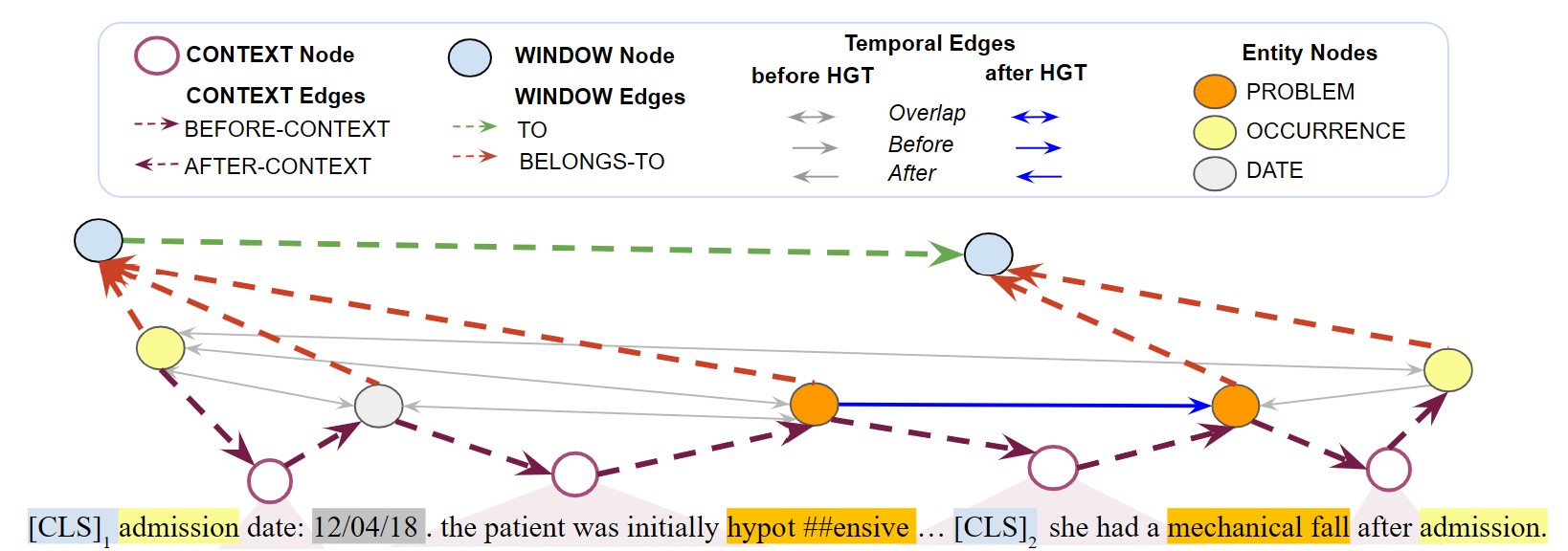}
    \caption{Heterogeneous Graph Construction. The gray edges represent intermediate high-confidence temporal relations inferred before HGT integration. The blue arrow represents additional relation inferred by the HGT component through local neighborhood aggregation via the entity and CONTEXT nodes and global neighborhood aggregation via the WINDOW nodes.}
     \vspace{-.1in}
    \label{fig:HGT}
\end{figure*}
The clinical texts are long documents with numerous entities and relations. The main motivation for using graphs is that these long documents contain implicit temporal dependencies that extend beyond local neighborhoods, making them difficult to capture with span-based approaches alone. 
To discover the global patterns, we create heterogeneous graphs from whole documents, with entities as nodes and their temporal relations as edges. We include special landmark nodes to incorporate local and global context. Our model \model works in three steps\textemdash (i) obtaining domain-aware representations, (ii) constructing a heterogeneous graph, and (iii) obtaining final predictions with \gnn.

\subsection{Domain-aware Representations (\spanmodel)}
We adopt a span-based approach to obtain domain-aware features, where a span is a contiguous sequence of tokens to be classified as an entity. We enumerate all spans up to length $k=7$, covering the $97^{th}$ percentile of entity lengths in the training data. Span representations are derived using BioMedBERT \citep{gu2021domain}, pre-trained from scratch on biomedical articles to capture domain-specific vocabulary. For example, BioMedBERT recognizes `creatinine' as a single token unlike general-domain BERT, which splits it into less meaningful subwords [`cr', `\#\#ea', `\#\#tin', `\#\#ine']. The representation $\mathbf{e}_{\text{sp}}$ of a span $sp$ is obtained by concatenating BioMedBERT embeddings $\rho$ for the span's start token ($\text{start[sp]}$), end token ($\text{end[sp]}$), and span-width embedding ($\omega(sp$) encoding the number of tokens in a span ($|sp|$).
\begin{equation}
\mathbf{e}_{\text{sp}} = FFNN(\rho_{\text{start[sp]}};
\rho_{\text{end[sp]}}; 
\omega(\text{sp}))
\end{equation}
\noindent A feed-forward network with two layers and ReLU activation is used as an entity decoder.
\begin{align}
\hat{y}^{n}_{i} &= \text{argmax}_{\ntypes_{sp}}(g(\mathbf{e}_{\text{sp}_i}))\\
g(\mathbf{e}_{\text{sp}_i}) &= \text{softmax}(FFNN(\mathbf{e}_{\text{sp}_i}))
\end{align}
where $\ntypes_{sp}$ is the set of entity/span types, including the NOT-ENTITY class for the remaining spans. 

For relation classification, we create pair-wise representations using entities not classified as NOT-ENTITY. We concatenate each entity span representation with the element-wise product of the two spans to capture the interactions between the spans. We also include the max-pooled representation of all the token embeddings between two entity spans. This design choice is inspired by several previous studies that observe significant performance improvement in relation extraction tasks by providing additional context between two entities \citep{cheng2023typed, li2023evaluating}--- i.e. $\mathbf{e}_{\text{ctx}}$ denoted the context between tokens $i$ and $j$. Additionally, we also provide the entity types as inferred by the entity decoder to the model for modeling domain-specific distributional constraints \citep{han2020domain, cheng2023typed}:
\begin{align}
\nonumber
&\hspace{-1em} \boldsymbol{\epsilon}_{i,j}=[\mathbf{e}_{\text{sp}_{i}}; \mathbf{e}_{\text{sp}_{j}};\mathbf{e}_{\text{sp}_{i}} \odot \mathbf{e}_{\text{sp}_{j}} ;\mathbf{e}_{\hat{y}^{n}_{i}};\mathbf{e}_{\hat{y}^{n}_{j}}; \mathbf{e}_{\text{ctx}(i,j)}] \\
& \hspace{-1em} \mathbf{e}_{\text{ctx}(i,j)}= \text{maxpool}(\\
\nonumber
&\{\rho_{k} \mid \text{end}[\text{sp}_i]< k < \text{start}[\text{sp}_j];\text{i}<\text{j}\})
\label{eq:epsilon_ij}
\end{align}
A relation decoder similar to the entity decoder is used to classify the pair representation of $i^{th}$ and $j^{th}$ spans into $\mathbf{Y}_\text{rtype}\in \text{\{Before, After, Overlap, NO-RELATION\}}$.
\begin{align}
\hat{y}^{r}_{ij} &= \text{argmax}_{\etypes_{sp}}(f(\boldsymbol{\epsilon}_{i,j}))\\
f(\boldsymbol{\epsilon}_{i,j}) &= \text{softmax}(FFNN(\boldsymbol{\epsilon}_{i,j}))
\label{eq:f_eq}
\end{align}
where $\etypes_{sp}$ is the set of relation types. 

For our task of joint end-to-end relation extraction, we use the combined training loss $\mathcal{L}= \mathcal{L}_{n} +\mathcal{L}_{r}$. Where $\mathcal{L}_{n}$ is the cross-entropy loss for entity and $\mathcal{L}_{r}$ for relation extraction.

\paragraph{Sliding Windows.}
Clinical notes are long, and BioMedBERT is limited to 512 tokens. Previous works often split the document into 1\textendash 3 sentence chunks, processing each separately, which results in loss of global context and potential inconsistencies. To overcome this, we use a sliding window approach to retain sufficient local context for each token's embedding \citep{beltagy2020longformer}. The sliding window mechanism processes a sequence by moving a fixed-size ($n$ tokens) window with a stride of $\lfloor\frac{n-2}{2} \rfloor$. The window masks ensure that despite overlapping windows to provide sufficient context to all tokens, the final representation of each token comes from a unique window (See Appendix \ref{app:slidingwin} for details). These local representations are then used for \trex.

We now describe our model \model that uses the predicted entities and their most confident Tlinks and refines the extracted temporal graph using graph neural networks (\gnn). 
 \begin{figure*}[!htbp]
  \centering
      {\includegraphics[width=\linewidth,trim=10 40 20 60,clip]{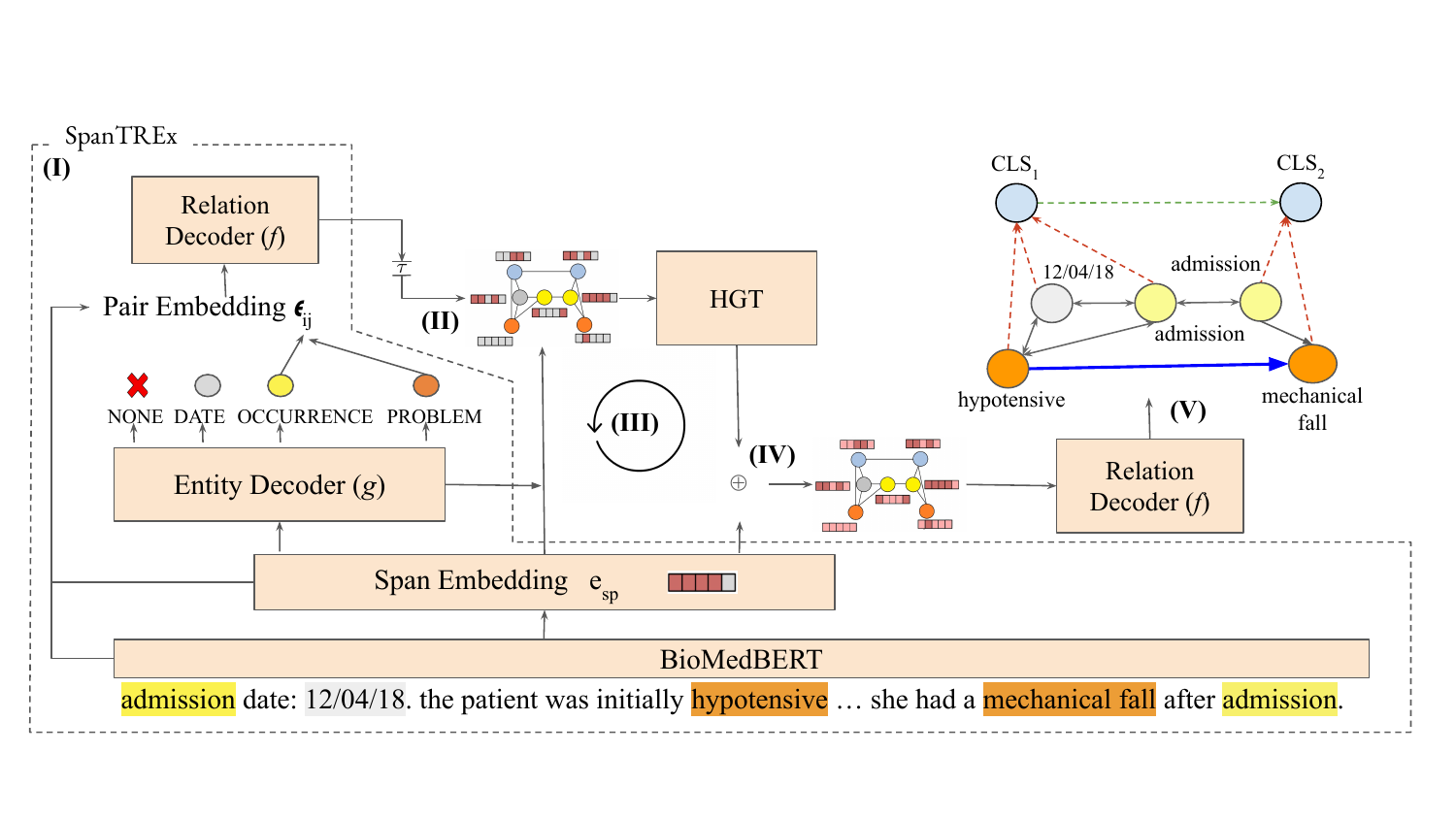}}
  \caption{\model Architecture. (I) The dashed boundary represents \spanmodel that produces the initial temporal graph. (II) A confidence threshold ($\tau$) filters uncertain edges (TLinks). (III) The edges are iteratively refined with the HGT component. (IV) The final node representations from the graph are combined with weighted residual of original span embeddings. (V) The relation decoder predicts the final temporal relations based on the HGT-enhanced representation. This way \model infers additional local and global relations (\textit{blue arrow}).}
   \vspace{-3mm}
  \label{fig:HGTSpanTREx}
\end{figure*}
\subsection{Document-Level Heterogeneous Graph}\label{sec:hg_construction}
A Heterogeneous Graph (HG) models heterogeneities in both the node and the edge types \citep{hetersurvey2017}. It is defined as a tuple $\graph=(\nodes, \edges, \ntypes, \etypes)$, where $\nodes = \{v_i\}_{i=1}^{N_v}$ and $\edges = \{e_j\}_{j=1}^{N_e}$ are the sets of nodes and edges. The node types and edge types are denoted by $\ntypes$ and $\etypes$, with a unique mapping for each node $\phi(v): \nodes \rightarrow \ntypes$ and edge $\psi(e): \edges \rightarrow \etypes$. Here $N_v$ and $N_e$ represent the total number of nodes and edges. We can model the graph as a set of triples $(\phi(s), \psi(e), \phi(t))$, or ``meta relations'', for every source node $s$ connected to target node $t$ by edge $e$. To construct an HG from the documents (illustrated in Figure \ref{fig:HGT}), we define two categories of node types, namely, \textit{Token-level} nodes (Entity/CONTEXT) and \textit{Document-level} nodes (WINDOW) to incorporate local and global context, allowing information exchange between both the neighboring entities and across the windows. We describe these in the following paragraphs and further elaborate in Appendix \ref{app:graph_details}.
\paragraph{Token-level Nodes.} 
Node types in this group capture the contextual understanding of the spans and their local inter-dependencies. These include \textit{Span-based} and \textit{Context-based} nodes.

\textit{Entity Nodes. }We represent every span that is not predicted as NOT-ENTITY by $g$ as a node in this subgroup. The node type for these is the predicted entity type. Thus, we obtain $
|\ntypes_{sp}|-1$ node types intitialized with their representation $e_{sp}$. We connect two such nodes if $f$ predicts their corresponding temporal relation other than NO-RELATION, with the prediction probability above a pre-defined threshold $\tau$. This \textbf{confidence thresholding} ensures that information is propagated only via the most certain relation predictions of $f$. This is particularly important in the early stages of training, as all components are trained simultaneously without certainty filtering, a sub-optimal model might propagate the noise across the graph. 

\textit{Context-based Nodes.} In natural language, the meaning of a word or phrase is closely tied to its local context \citep{gpt12018, bert2019, mikolov2013distributed}. We also observe this in our experiments with \spanmodel, where incorporating additional context between spans brings a high-performance gain in relation extraction (see Figure \ref{fig:pair_representation}). To capture this in the HG, we introduce \textit{context nodes} that store pooled context between a span pair. Directed edges connect the spans to the context node in the lexical order. Introducing context nodes for all possible span pairs results in a dense graph where even far-apart spans are reachable within 1- or 2-hops, leading to the common over-smoothing issue after information aggregation. To mitigate this, we limit context nodes to span pairs within a distance of $\delta$ tokens.
\paragraph{Document-level Nodes.}
To detect the long-range dependencies, we introduce window nodes which act as global landmarks in the document. Given a common window length $L_w$ for all documents, there can be $\left\lceil \frac{L}{L_w} \right\rceil$ window nodes in a document of length $L$ while the node type WINDOW is assigned to all of them. Entity nodes within a window connect to their window node via distinct edge types based on the entity types, and window nodes are connected sequentially based on their lexical order. Window nodes are of significant importance since they help aggregate information from far neighborhoods and help in finding transitive relations. 
\subsection{Final Predictions by \model}
Once an HG is constructed, we model it with the heterogeneous graph transformer (HGT) \citep{hgt2020}. Unlike traditional \gnn typically applied to homogeneous graphs \cite{gat2018}, HGT considers node- and edge-type distributions. The HGT module takes a document graph $\mathcal{G}$ constructed as discussed in Section \ref{sec:hg_construction} and initializes entity nodes with span embeddings ($\mathbf{e}_{\text{sp}}$). To produce structure-enhanced representations, the HGT module applies a message-passing process and enriches span embeddings with the aggregated information from their neighborhoods in the graph to extract global relation patterns effectively (the HGT module is described in Appendix \ref{app:hgt_module}). 

Figure \ref{fig:HGTSpanTREx} shows the complete \model architecture. After obtaining HGT representations for entities, we add them to the weighted residual of the original span embeddings. The new representations are then fed to the relation decoder. Here the nodes with \textsc{CLS} labels refer to the Window nodes (we do not show the context nodes for clarity). The example highlights that while \spanmodel can infer the \textit{Overlap} ($=$) relation between co-referring mentions of ``admission'', HGT integration in \model allows us to go further. For example, by aggregating information that `hypotensive' $=$ `admission' and `mechanical fall' $>$ `admission', HGT helps detect the long-range relation `hypotensive' $<$ `mechanical fall' as shown with the blue arrow.
\section{Experiments}
\label{sec:experiments}
\begin{table*}[htbp]
    \centering
     \adjustbox{max width=.95\textwidth}{
    \begin{tabular}{l c c c c c c c}
\toprule
 \multirow{3}{*}{\textbf{Model}} 
 & \multicolumn{2}{c}{\textbf{EVENT} }
 &\multicolumn{2}{c}{\textbf{TimEx}}
 &\multicolumn{3}{c}{\textbf{TLink}}\\
 \cmidrule(r){2-8}
 & \textbf{EI}&\textbf{EC} & \textbf{EI}&\textbf{EC}&\multicolumn{3}{c}{\textbf{RE}}\\
  \cmidrule(r){2-8}&$F_{1}$&Acc&$F_{1}$&Acc&P&R&F$_{1}$\\
\midrule 
Rule+ML \citep{xu2013end} 
        &  \textbf{91.66} & \textbf{85.74}
        &\textbf{91.44} &\textbf{89.29}
        & 67.10 & 60.01 & 63.36\\
Rule+ML \citep{tang2013hybrid}  &90.13 &83.60 & 
             86.59 & 85.00  
             &70.06	&56.88 & 62.78 \\
 REBEL+BART \citep{saiz2023end}& 78.00 & 72.00 
                            & 77.00 & 65.00
                            & 65.00 & 52.00 & 58.00 \\  
Multi-head Attention \cite{miller2023end}&89.42&80.74&89.01&77.42&86.12&29.76&44.24\\
SPERT \citep{eberts2020span}&89.44	&81.1			&90.29&	82.64	&77.58	&50.36&	61.08\\
 \spanmodel (ours) & 89.49& 81.35
            & 90.13& 81.32 
            &\textbf{81.33}& 56.44 & 66.63\\
                          
  \model (ours) & 89.55& 80.99
                & 90.06& 81.21
                &78.21 & \textbf{61.42} & \textbf{68.81}\\
\bottomrule
\end{tabular}
}
\caption{I2B2 2012 end-to-end entity-relation extraction task results using TLink $F_1$ as the primary metric.}
\vspace{-.5cm}
    \label{tab:end2end}
\end{table*}

\subsection{Reproducibility} The implementation details are in Appendix \ref{app:implementation}, and our code is available at \url{https://github.com/pbaghershahi/GraphTREX}. 
\subsection{Evaluation Setup}
The event extraction and timex extraction tasks in the I2B2 2012 challenge are evaluated using span $F_1$ (EI for entity identification) rewarding partial span match, and entity type accuracy (EC for entity classification). For the end-to-end relation extraction (RE), the TLink $F_1$ is considered as the primary metric. Following the I2B2 2012 shared task, we exclude the NO-RELATION pairs for final evaluation. An important consideration while evaluating Temporal Relation Extraction (\trex)  methods is that the same temporal order of events can be defined in multiple ways. For example, (A$<$B, B=C) can also be specified as (A$<$C, B=C) due to transitivity, and (A$<$B) is the same as (B$>$A) due to invertibility. Therefore, the temporal awareness metrics \citep{uzzaman2011temporal} are adopted for evaluation:
\begin{equation*}
     \scalebox{0.9}{$P = \frac{|G^{-}_{sys} \cap G^{+}_{gold}|}{G^{-}_{sys}}, \quad R = \frac{|G^{-}_{gold} \cap G^{+}_{sys}|}{G^{-}_{gold}}, \quad F_{1}=2 \frac{PR}{P+R}$}
\end{equation*}
where $G^{+}$ is the closure of graph $G$ that makes implicit temporal relations explicit by reasoning on a set of relations using transitivity and invertibility rules. $G^{-}$ is the reduced graph after removing redundant relations inferable from other relations of $G$. The intersections in the formulae refer to the TLinks of the corresponding graphs.
For E3C corpus, we use micro-F1 scores for both entity and relations following the THYME standard.
\paragraph{Baselines.} For the I2B2 corpus, we evaluate \model and \spanmodel against hybrid rule and ML-based approaches \citep{xu2013end, tang2013hybrid}, a recent generative framework \citep{saiz2023end}, popular general-domain model SPERT \cite{eberts2020span}, and multi-head attention-based method having state-of-the-art on the THYME corpus \citet{miller2023end}.\footnote{To ensure a fair comparison, we use the same entity-extraction setup as \spanmodel and implement multi-head attention over an $n\times n$ matrix of entities for relation decoder.}. We also experiment with a pipeline approach\textemdash extracting entities using our state-of-the-art BioMedBERT-UMLS model (Appendix \ref{app:additionalencoders}) and classifying entity pairs using the method from \citet{zhong2021frustratingly}, which attains state-of-the-art in I2B2 2012 relation classification \citep{cheng2023typed}. We extend this to include the majority NO-RELATION class for pairs without clear temporal links. Similar to earlier works, we are unable to run this model for document-level pairs due to computational constraints and limit this experiment to same-sentence pairs (see Appendix \ref{app:pipeline}). For E3C, we compare \spanmodel and \model against multi-head attention and SPERT. \footnote{We exclude baselines from 2013 due to code unavailability and additional private annotations, and REBEL and pipeline approaches due to poor performance on I2B2 2012.}

\subsection{Results and Discussion}
Table \ref{tab:end2end} shows that \model sets a new state-of-the-art in end-to-end \trex on I2B2 2012, with a 68.81\% tempeval $F_1$, a 5.45\% gain over \citet{xu2013end}. \model also outperforms \spanmodel with more than 2\% improvement due to an improved recall. The improvement in document-level $F_1$ scores is also highly statistically significant ($\text{p-value} = 4.26  \times 10^{-9}$, paired t-test). Even with added model complexity, these gains are crucial for the clinical domain and patient outcomes. The multi-head attention method has the lowest performance among all baselines despite achieving state-of-the-art on the THYME corpus. While multi-head attention between two entities can effectively capture word-level dependencies, it may still struggle with long-range dependencies without additional global or local context. In contrast, \spanmodel incorporates additional local context. \model further brings node-specific and edge-specific multi-head attention and provides additional global context, allowing interactions between existing high-confidence relations to improve the inference over long-range relations too. Additionally, the pipeline approach, limited to same-sentence relations, struggles with the NO-RELATION class, achieving only 10.79\% $F_1$. In contrast, \model effectively distinguishes relevant relations from the majority of negative samples, achieving 59.67\% $F_1$ in sentence-level pairs. We also train our model for relation extraction with gold entities and find that the model severely underperforms, showing the importance of joint training with shared representations. 

\paragraph{Qualitative Analysis} We conduct an analysis of the system outputs against the ground truth and present a case study in Appendix~\ref{app:errors}, Figure~\ref{fig:visualization}. The example reveals an inconsistency in the ground truth, where `discharge' event overlaps with both the admission and discharge dates. While \spanmodel replicates this error, \model does not, showing greater robustness. \model output is also more balanced across relation types and has denser connections than the ground truth, while \spanmodel identifies fewer \textit{Before}/\textit{After} pairs. \spanmodel misses two important wellness indicators\textemdash `tolerated' and `treatment'. Beyond this case study, we also find that the reported entity extraction scores may not fully reflect our model’s capability, as it identifies meaningful entities even when the gold standard omits them. For example, given \{admission', hospitalization'\}, \model and \spanmodel correctly identify both while the ground truth labels only the first.

\paragraph{Generalization to E3C Corpus}
We consider EVENTs and their TLinks since E3C data is sparsely annotated and majority (85.63\%) of annotated TLinks occur between EVENTs only (Appendix \ref{app:E3C}). 
Table \ref{tab:e3c} shows \spanmodel outperforms SPERT by a large margin and \model gives further 1\% improvement, showing strong generalizability of the method on this challenging dataset. We also attain improved entity extraction scores compared to the previous best of 63.44\% \citep{zanoli2024assessment}. 
\begin{table}[htbp]
    \centering
     \adjustbox{max width=.5\textwidth}{
    \begin{tabular}{l c c c c}
\toprule
 \multirow{2}{*}{\textbf{Model}} 
 & \textbf{EVENT}  &\textbf{TLink}\\
 \cmidrule(r){2-3}
 & $F_{1}$&$F_{1}$\\
\midrule
Multi-head Attention&84.0&3.97\\
SPERT &	78.85 &	13.63\\
\spanmodel (ours) &	81.3 &22.55 \\
\model (ours) &\textbf{82.10}&\textbf{23.48}\\
\bottomrule
\end{tabular}
}
\caption{Results on the E3C Corpus.}
    \label{tab:e3c}
\end{table}
 \vspace{-.6cm}
\subsection{Ablation Analyses}
Here we present further detailed analysis and ablation studies on the I2B2 corpus and provide additional experiments in Appendix \ref{app:additional_exp}.


\paragraph{Does the \gnn module improve long-range predictions?}
\begin{table}[htbp]
    \centering
    \adjustbox{max width=.5\textwidth}{
    \begin{tabular}{c c c c c c c c c}
\toprule
 \textbf{Distance }&\multirow{2}{*}{$\mathbf{\bar{n}_r}$}& \multicolumn{3}{c}{\textbf{\spanmodel}}&\multicolumn{3}{c}{\textbf{\model}} & \multirow{2}{*}{\textbf{\%IMP}}\\
 \cmidrule(r){3-8}
 ($d_r$)&&P &R & F$_{1}$&P &R & F$_{1}$&\\	
\midrule
 $d_r=0$ &159.4&  75.4 &41.1 &53.2&
                    72.4&45.1&55.6&4.5\\
$d_r> 0$  &70.6& 93.6 &15.6& 26.8 
 &91.0& 17.0 & 28.6&6.7 \\
 $d_r>1$ & 35.9& 94.4 & 8.6& 15.7& 92.8& 9.4&17.1&8.9\\
 \bottomrule
\end{tabular}
}
\caption{TLink scores by entity pair distance on I2B2: same window ($d=0$), across windows ($d>0, d>1$). $\mathbf{\bar{n}_r}$ is the average ground truth TLinks. \%IMP measures $F_1$ gain of \model over \spanmodel.
}
    \label{tab:distance_analysis}
    \vspace{-3mm}
\end{table}
We investigate whether local versus global information exchange across an HG improves performance for lexically distant entities. We compare the TLink predictions over the same context window ($d_r=0$) for local TLinks, across different windows ($d_r>0$), and farther than adjacent windows ($d_r>1$) for global TLinks. Here $d_r$ denotes the number of windows between two entities. Table \ref{tab:distance_analysis} shows \model outperforms \spanmodel across all distances, with a \% improvement (lift) of 4.5\% for same-window relations, 6.7\% for those spanning one or more windows, and 8.9\% for more distant pairs.

An important finding of this experiment is the \textit{scalability} potential of \model as it improves structural dependencies across long documents with graphs while keeping the relations and context invariant to the document length. This is an important advantage since the longer a document gets, the harder it is for a text-based model or transformer LMs to process it and capture long-range dependencies \citep{yuan-etal-2023-zero, qin-etal-2023-nlp}.

\begin{figure}[htbp]
\centering
    \includegraphics[
    trim=8 10 7 5, clip, 
    width=1\columnwidth, 
    ]{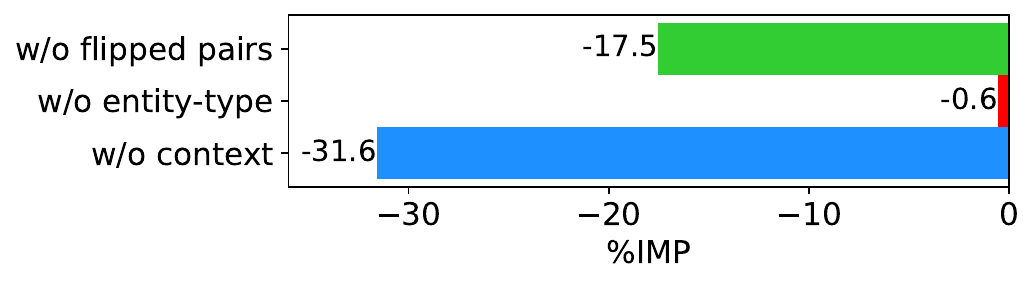}
    \caption{\% $F_1$ gain on I2B2 of \spanmodel over ablations omitting specific pair representation components.
    } \label{fig:pair_representation}
    \vspace{-5mm}
\end{figure}

\paragraph{Variations in Pair Embedding Approaches.} To enhance \spanmodel, we include (1) additional context by pooling all the tokens between entity spans and (2) predicted entity types in the pair embeddings ($\boldsymbol{\epsilon}_{\text{i,j}}$ ( Eq. \ref{eq:epsilon_ij}). (3) We also augment training data by adding inverse flipped relations\textemdash if $A=B$, we add $B=A$, if $A<B$, we add $B>A$ and vice-versa. Figure \ref{fig:pair_representation} shows the percentage improvement w.r.t. the \spanmodel, with the highest boost from additional context. Additional flipped relations also provide considerable gains while adding entity types leads to marginal gains. Even on the E3C corpus, additional context is crucial, as performance drops to 8.44\% without it, and without flipping the \textit{Overlap} relation, it drops to 20.95\%.\footnote{\textit{Before} pairs are not flipped in E3C since it doesn't have the inverse \textit{After} class.}

\paragraph{Ablation on Graph Construction/Impact of Node Types.}
We analyze the impact of each HG component by varying the node types used for graph construction and compare the performance with \spanmodel (Table \ref{tab:graphcomponents}). Using only entity nodes can degrade performance, likely due to the entity nodes overfitting their local neighborhoods in the absence of a broader context. Adding context nodes improves results,  consistent with Figure \ref{fig:pair_representation}, showing the importance of context between spans. Including both the context nodes and the window nodes further enhances the model performance, showing that they are mutually compatible.\footnote{Ablation on E3C corpus gives slightly different trends, given the short documents (Appendix \ref{app:ablatione3c}).}

\begin{table}[htbp]
    \small
    \centering
       \scalebox{0.8}{ 
       \begin{tabular}{L{0.6in}*{3}{C{0.25in}}*{3}{C{0.25in}}C{0.35in}}
            \toprule
            \multirow{2}{6em}{\textbf{Method}} & \multicolumn{3}{c}{\textbf{Graph Components}} & & & & \\
            \cmidrule(lr){2-4}
            & \rotatebox{90}{\parbox{1cm}{\centering \textbf{Span}}} & \rotatebox{90}{\parbox{1.1cm}{\centering \textbf{Context}}} & \rotatebox{90}{\parbox{1.1cm}{\centering \textbf{Window}}}& \textbf{P} & \textbf{R} & \textbf{F} & \textbf{\%IMP} \\
            \midrule
            \spanmodel & & & & \textbf{81.33} & 56.44 & 66.63 & - \\
            \midrule
            \multirow{4}{6em}{\model} & \ding{51} & \ding{55} & \ding{55} & 80.26 & 53.67 & 64.43 & -3.3 \\
            & \ding{51} & \ding{51} & \ding{55} & 80.34 & 58.75 & 67.87 & 1.9 \\
            & \ding{51} & \ding{55} & \ding{51} & 77.23 & 59.98 & 67.52 & 1.3 \\
            & \ding{51} & \ding{51} & \ding{51} & 78.21 & \textbf{61.42} & \textbf{68.81} & \textbf{2.3} \\
            \bottomrule
        \end{tabular}
        }
                \caption{\% $F_1$ gains (IMP) of \model variants over \spanmodel for the I2B2 dataset.}
    \label{tab:graphcomponents}    	
\end{table}

\paragraph{Variations Across Different Encoders}
We experiment with several other base encoders and report the results with two popular models\textemdash BERT and RoBERTa\textemdash in Table \ref{tab:additional_encoders}. We find that \model provides additional gains over \spanmodel across both encoders. We also report the results with clinical Longformer \cite{li2022clinical}, designed to process long clinical texts. In line with \citet{qin2023nlp} we find that in their native implementation using \spanmodel, the transformer-based models including Longformer struggle with long-range dependencies.  Clinical Longformer also underperforms over \model with $F_1$ score  63.84\%. We also report the results on entity extraction with several pre-trained clinical base encoders in Appendix \ref{app:additionalencoders} Table \ref{tab:clinical_encoders}, and find that none of them outperform BioMedBERT.
\begin{table}[htbp]
    \centering
        \adjustbox{max width=.5\textwidth}{
    \begin{tabular}{l c c c}
\toprule
 \multirow{2}{*}{\textbf{Model}} 
 &\multicolumn{3}{c}{\textbf{TLink (end-to-end)}}\\
 &P&R&F$_{1}$\\
\midrule              
  \model        &78.21 & \textbf{61.42} & \textbf{68.81}\\

\midrule
BERT-\spanmodel& 75.94&	48.63	&59.30\\
BERT-\model &77.45	&52.30	&62.44\\
RoBERTA-\spanmodel & 79.98&	52.05&	63.06\\
RoBERTa-\model&	78.22&	54.88&	64.50\\
\midrule
Clinical-Longformer&\textbf{82.39}&52.10&63.84\\
\bottomrule
\end{tabular}
}
    \caption{Variations across encoders on the I2B2 data.}
    \label{tab:additional_encoders}
\end{table}
\vspace{-.5cm}

\paragraph{Performance Across Relation Classes}
We report the performance of \model across individual TLink classes in Table \ref{tab:classwise}. While reporting these scores for E3C is straightforward since the evaluation metric is micro $F_1$, for I2B2, we report tempeval metrics. Therefore, before computing these metrics, we filter the system and gold output to the desired relation type. Note that during training, we enrich the training data with corresponding inverse relations by flipping the participant entity order, due to which the model learns to output an equivalent number of \textit{Before}/\textit{After} relations. However, for the gold standard, there are fewer TLinks for \textit{After}. Therefore, to have a fair evaluation, we similarly flip the relation type in the gold standard before computing the above scores. The number of TLinks reported in the above table is based on the I2B2 evaluation script; they represent the total compatible links in the gold standard test set, given the match in extracted entities. As seen from this table, the model performance is better on the majority class for both datasets. 
\begin{table}[ht]
    \centering
    \adjustbox{max width=.45\textwidth}{
    \begin{tabular}{lccc}
\toprule
Corpus	&RelType&	\#TLinks	&F1\\
\midrule
\multirow{3}{*}{I2B2 2012}	&\textit{Overlap}&	7155	&59.27\\
&\textit{Before}&	13453	&75.05\\
&\textit{After}&	13453&	74.94\\
\midrule
\multirow{2}{*}{E3C}	&\textit{Overlap}&	1560	&24.23\\
&\textit{Before}	&793	&21.88\\
\bottomrule
\end{tabular}
}
    \caption{Performance Across Relation Classes.}
    \label{tab:classwise}
\end{table}
\vspace{-6mm}

\section{Conclusion and Future Work}
We introduce \model, a novel model for temporal relation extraction in long clinical documents. By integrating span-based extraction with GNNs, \model provides a structured and context-aware framework, achieving a new state-of-the-art in end-to-end \trex on the I2B2 2012 corpus. Our ablation studies show that constructing a document-level heterogeneous graph with special landmark nodes more effectively captures local and global information, enhancing long-range relation extraction. Experiments on the E3C corpus demonstrate the model's generalizability beyond the U.S. context. We also introduce BioMedBERT-UMLS, a knowledge-infused entity extraction model achieving state-of-the-art in TimEx extraction. Due to non-trivial scalability challenges, we consider knowledge integration for \trex a future direction.

Accurate event chronology is crucial for clinical applications, including opportunistic screening for chronic diseases like Type 2 Diabetes, which affects millions worldwide \cite{IDF2025} and is a key focus of our ongoing research. This is especially vital for marginalized communities lacking regular access to care. Clinical notes from even a single patient visit capture richer historical context than structured EHR but remain underutilized. By enabling more precise \trex from these narratives, \model lays the foundation for enhanced event sequence modeling for health risk prediction, treatment optimization, clinical decision support, and patient education. Integrating such systems with human-in-the-loop approaches can mitigate missed diagnoses,  facilitate opportunistic screening, and advance healthcare for all. 

\section*{Limitations}
One of the limitations of the current work is the constraint to validate the model across more clinical datasets due to their unavailability. In the clinical domain, concerns about the leakage of protected health information limit data release and cross-hospital annotation efforts. The only other moderate-scale dataset THYME \cite{bethard-etal-2017-semeval, styler2014temporal},  requires a membership fee similar to the Linguistic Data Consortium (LDC). However, we show generalizability to another recent small-scaled dataset. Due to the ease of implementation, code availability, and adaptability of our approach, it can be extended for the future works once additional datasets become available or accessible. 

More broadly, \model enables downstream research in healthcare and clinical decision support, rather than as a standalone diagnostic tool. In real-world applications, clinical NLP models like ours are typically integrated into a broader analytics ecosystem, for example, in Type 2 Diabetes opportunistic screening research in our own research ecosystem, where we are applying it to multiple note types beyond just discharge summaries and applying it for multi-document temporal reasoning. While strong model performance has positive societal impacts\textemdash such as improving early risk detection, optimizing treatment pathways, reducing clinician burden, and improving patient education with summarized timelines\textemdash these systems should be used with human oversight to manage the risks associated with model limitations. Future research should focus on robust evaluation across diverse clinical settings, to ensure reliable, and contextually appropriate deployment of the system.

\section*{Ethics Statement}
This research uses de-identified clinical notes from Beth Israel Medical Center in Boston. The data was released by Harvard Medical School and is available with necessary ethical approvals. We have signed a Data Use Agreement (DUA) with n2c2 to ensure compliance with privacy regulations and safeguard data confidentiality. We take all necessary precautions to prevent misuse of the data and unauthorized access to the data. Additionally, we also show experimental results on the E3C corpus constructed from publicly available clinical documents. Our work aims to enhance clinical decision-making by extracting temporal relations from medical events, supporting healthcare professionals' judgment without undermining it. By accurately extracting detailed information from clinical notes, which capture the nuances of patient history, \model can provide a more comprehensive and accurate summary of patient timelines. This capability helps bridge diagnostic disparities, particularly for marginalized communities with irregular access to care or incomplete medical records, ensuring healthcare providers have a more complete view of a patient's medical journey. Given the potential impact on patient outcomes, ongoing validation and evaluation are essential to ensure that the model’s performance is robust, accurate, and equitable, mitigating risks of misinterpretation and unintended harm.  

\section*{Acknowledgment}
We thank the anonymous reviewers for their valuable feedback. This work is supported by NSF IIS \#2312862.
\clearpage
\bibliography{acl2025}

\begin{thebibliography}{85}
\providecommand{\natexlab}[1]{#1}

\bibitem[{IDF(2025)}]{IDF2025}
 2025.
\newblock \href {https://diabetesatlas.org/resources/idf-diabetes-atlas-2025/} {\emph{IDF Diabetes Atlas}}, 11 edition.
\newblock International Diabetes Federation, Brussels, Belgium.

\bibitem[{Alfattni et~al.(2020)Alfattni, Peek, and Nenadic}]{alfattni2020extraction}
Ghada Alfattni, Niels Peek, and Goran Nenadic. 2020.
\newblock Extraction of temporal relations from clinical free text: A systematic review of current approaches.
\newblock \emph{Journal of biomedical informatics}, 108:103488.

\bibitem[{Allen(1983)}]{allen1983maintaining}
James~F Allen. 1983.
\newblock Maintaining knowledge about temporal intervals.
\newblock \emph{Communications of the ACM}, 26(11):832--843.

\bibitem[{Alsentzer et~al.(2019)Alsentzer, Murphy, Boag, Weng, Jin, Naumann, Redmond, and McDermott}]{alsentzer2019publicly}
Emily Alsentzer, John~R Murphy, Willie Boag, Wei-Hung Weng, Di~Jin, Tristan Naumann, WA~Redmond, and Matthew~BA McDermott. 2019.
\newblock Publicly available clinical bert embeddings.
\newblock \emph{NAACL HLT 2019}, page~72.

\bibitem[{Aronson et~al.(1994)Aronson, Rindflesch, and Browne}]{aronson1994exploiting}
Alan~R Aronson, Thomas~C Rindflesch, and Allen~C Browne. 1994.
\newblock Exploiting a large thesaurus for information retrieval.
\newblock In \emph{RIAO}, volume~94, pages 197--216.

\bibitem[{Baghershahi et~al.(2023)Baghershahi, Hosseini, and Moradi}]{efficientrelation2023}
Peyman Baghershahi, Reshad Hosseini, and Hadi Moradi. 2023.
\newblock \href {https://arxiv.org/abs/2212.05581} {Efficient relation-aware neighborhood aggregation in graph neural networks via tensor decomposition}.
\newblock \emph{Preprint}, arXiv:2212.05581.

\bibitem[{Beltagy et~al.(2020)Beltagy, Peters, and Cohan}]{beltagy2020longformer}
Iz~Beltagy, Matthew~E Peters, and Arman Cohan. 2020.
\newblock Longformer: The long-document transformer.
\newblock \emph{arXiv preprint arXiv:2004.05150}.

\bibitem[{Bethard et~al.(2017)Bethard, Savova, Palmer, and Pustejovsky}]{bethard-etal-2017-semeval}
Steven Bethard, Guergana Savova, Martha Palmer, and James Pustejovsky. 2017.
\newblock \href {https://doi.org/10.18653/v1/S17-2093} {{S}em{E}val-2017 task 12: Clinical {T}emp{E}val}.
\newblock In \emph{Proceedings of the 11th International Workshop on Semantic Evaluation ({S}em{E}val-2017)}, pages 565--572, Vancouver, Canada. Association for Computational Linguistics.

\bibitem[{Bodenreider(2004)}]{bodenreider2004unified}
Olivier Bodenreider. 2004.
\newblock The unified medical language system (umls): integrating biomedical terminology.
\newblock \emph{Nucleic acids research}, 32(suppl\_1):D267--D270.

\bibitem[{Capurro et~al.(2014)Capurro, van Eaton, Black, and Tarczy-Hornoch}]{capurro2014availability}
Daniel Capurro, Erik van Eaton, Robert Black, and Peter Tarczy-Hornoch. 2014.
\newblock Availability of structured and unstructured clinical data for comparative effectiveness research and quality improvement: a multisite assessment.
\newblock \emph{EGEMS}, 2(1).

\bibitem[{Chan et~al.(2023)Chan, Cheng, Wang, Jiang, Fang, Liu, and Song}]{chan2023chatgpt}
Chunkit Chan, Jiayang Cheng, Weiqi Wang, Yuxin Jiang, Tianqing Fang, Xin Liu, and Yangqiu Song. 2023.
\newblock Chatgpt evaluation on sentence level relations: A focus on temporal, causal, and discourse relations.
\newblock \emph{arXiv preprint arXiv:2304.14827}.

\bibitem[{Chaturvedi et~al.(2023)Chaturvedi, Rashid, Layden, Boyd, Cinar, and Di~Eugenio}]{chaturvedi2023sequential}
Rochana Chaturvedi, Mudassir Rashid, Brian~T Layden, Andrew Boyd, Ali Cinar, and Barbara Di~Eugenio. 2023.
\newblock Sequential representation of sparse heterogeneous data for diabetes risk prediction.
\newblock In \emph{2023 IEEE International Conference on Bioinformatics and Biomedicine (BIBM)}, pages 831--834. IEEE.

\bibitem[{Chen et~al.(2021)Chen, Wang, and Wang}]{chen2dataset}
Wenhu Chen, Xinyi Wang, and William~Yang Wang. 2021.
\newblock A dataset for answering time-sensitive questions.
\newblock In \emph{Thirty-fifth Conference on Neural Information Processing Systems Datasets and Benchmarks Track (Round 2)}.

\bibitem[{Cheng and Weiss(2023)}]{cheng2023typed}
Cheng Cheng and Jeremy~C Weiss. 2023.
\newblock Typed markers and context for clinical temporal relation extraction.
\newblock In \emph{Machine Learning for Healthcare Conference}, pages 94--109. PMLR.

\bibitem[{Cui et~al.(2023)Cui, Han, and Nenadic}]{cui2023medtem2}
Yang Cui, Lifeng Han, and Goran Nenadic. 2023.
\newblock Medtem2. 0: Prompt-based temporal classification of treatment events from discharge summaries.
\newblock In \emph{Proceedings of the 61st Annual Meeting of the Association for Computational Linguistics (Volume 4: Student Research Workshop)}, pages 160--183.

\bibitem[{Devlin et~al.(2019)Devlin, Chang, Lee, and Toutanova}]{bert2019}
Jacob Devlin, Ming-Wei Chang, Kenton Lee, and Kristina Toutanova. 2019.
\newblock {BERT}: Pre-training of deep bidirectional transformers for language understanding.
\newblock In \emph{Proceedings of the 2019 Conference of the North {A}merican Chapter of the Association for Computational Linguistics: Human Language Technologies, Volume 1 (Long and Short Papers)}, pages 4171--4186.

\bibitem[{Dhippayom et~al.(2013)Dhippayom, Fuangchan, Tunpichart, and Chaiyakunapruk}]{dhippayom2013}
Teerapon Dhippayom, Anjana Fuangchan, Sirirat Tunpichart, and Nathorn Chaiyakunapruk. 2013.
\newblock Opportunistic screening and health promotion for type 2 diabetes: an expanding public health role for the community pharmacist.
\newblock \emph{Journal of Public health}, 35(2):262--269.

\bibitem[{Dixit and Al-Onaizan(2019)}]{dixit2019span}
Kalpit Dixit and Yaser Al-Onaizan. 2019.
\newblock Span-level model for relation extraction.
\newblock In \emph{Proceedings of the 57th Annual Meeting of the Association for Computational Linguistics}, pages 5308--5314.

\bibitem[{Dligach et~al.(2017)Dligach, Miller, Lin, Bethard, and Savova}]{dligach2017neural}
Dmitriy Dligach, Timothy Miller, Chen Lin, Steven Bethard, and Guergana Savova. 2017.
\newblock Neural temporal relation extraction.
\newblock In \emph{Proceedings of the 15th Conference of the European Chapter of the Association for Computational Linguistics: Volume 2, Short Papers}, pages 746--751.

\bibitem[{Eberts and Ulges(2020)}]{eberts2020span}
Markus Eberts and Adrian Ulges. 2020.
\newblock Span-based joint entity and relation extraction with transformer pre-training.
\newblock In \emph{ECAI 2020}, pages 2006--2013. IOS Press.

\bibitem[{Fu et~al.(2020)Fu, Liu, and Neubig}]{fu2020interpretable}
Jinlan Fu, Pengfei Liu, and Graham Neubig. 2020.
\newblock Interpretable multi-dataset evaluation for named entity recognition.
\newblock In \emph{Proceedings of the 2020 Conference on Empirical Methods in Natural Language Processing (EMNLP)}, pages 6058--6069.

\bibitem[{Gao et~al.(2023)Gao, Zhao, Yu, and Xu}]{gao2023exploring}
Jun Gao, Huan Zhao, Changlong Yu, and Ruifeng Xu. 2023.
\newblock Exploring the feasibility of chatgpt for event extraction.
\newblock \emph{arXiv preprint arXiv:2303.03836}.

\bibitem[{Ghassemi et~al.(2015)Ghassemi, Pimentel, Naumann, Brennan, Clifton, Szolovits, and Feng}]{ghassemi2015multivariate}
Marzyeh Ghassemi, Marco A.~F. Pimentel, Tristan Naumann, Thomas Brennan, David~A. Clifton, Peter Szolovits, and Mengling Feng. 2015.
\newblock A multivariate timeseries modeling approach to severity of illness assessment and forecasting in icu with sparse, heterogeneous clinical data.
\newblock In \emph{Proceedings of the Twenty-Ninth AAAI Conference on Artificial Intelligence}, AAAI'15, page 446–453. AAAI Press.

\bibitem[{Gu et~al.(2021)Gu, Tinn, Cheng, Lucas, Usuyama, Liu, Naumann, Gao, and Poon}]{gu2021domain}
Yu~Gu, Robert Tinn, Hao Cheng, Michael Lucas, Naoto Usuyama, Xiaodong Liu, Tristan Naumann, Jianfeng Gao, and Hoifung Poon. 2021.
\newblock Domain-specific language model pretraining for biomedical natural language processing.
\newblock \emph{ACM Transactions on Computing for Healthcare (HEALTH)}, 3(1):1--23.

\bibitem[{Han et~al.(2019)Han, Ning, and Peng}]{han2019joint}
Rujun Han, Qiang Ning, and Nanyun Peng. 2019.
\newblock Joint event and temporal relation extraction with shared representations and structured prediction.
\newblock In \emph{Proceedings of the 2019 Conference on Empirical Methods in Natural Language Processing and the 9th International Joint Conference on Natural Language Processing (EMNLP-IJCNLP)}, pages 434--444.

\bibitem[{Han et~al.(2020)Han, Zhou, and Peng}]{han2020domain}
Rujun Han, Yichao Zhou, and Nanyun Peng. 2020.
\newblock Domain knowledge empowered structured neural net for end-to-end event temporal relation extraction.
\newblock In \emph{Proceedings of the 2020 Conference on Empirical Methods in Natural Language Processing (EMNLP)}, pages 5717--5729.

\bibitem[{Hersh et~al.(2013)Hersh, Weiner, Embi, Logan, Payne, Bernstam, Lehmann, Hripcsak, Hartzog, Cimino et~al.}]{hersh2013caveats}
William~R Hersh, Mark~G Weiner, Peter~J Embi, Judith~R Logan, Philip~RO Payne, Elmer~V Bernstam, Harold~P Lehmann, George Hripcsak, Timothy~H Hartzog, James~J Cimino, et~al. 2013.
\newblock Caveats for the use of operational electronic health record data in comparative effectiveness research.
\newblock \emph{Medical care}, 51:S30--S37.

\bibitem[{Hu et~al.(2020)Hu, Dong, Wang, and Sun}]{hgt2020}
Ziniu Hu, Yuxiao Dong, Kuansan Wang, and Yizhou Sun. 2020.
\newblock Heterogeneous graph transformer.
\newblock In \emph{Proceedings of The Web Conference 2020}, page 2704–2710.

\bibitem[{Huguet~Cabot and Navigli(2021)}]{huguet-cabot-navigli-2021-rebel-relation}
Pere-Llu{\'\i}s Huguet~Cabot and Roberto Navigli. 2021.
\newblock \href {https://doi.org/10.18653/v1/2021.findings-emnlp.204} {{REBEL}: Relation extraction by end-to-end language generation}.
\newblock In \emph{Findings of the Association for Computational Linguistics: EMNLP 2021}, pages 2370--2381, Punta Cana, Dominican Republic. Association for Computational Linguistics.

\bibitem[{Lai et~al.(2021)Lai, Ji, Zhai, and Tran}]{lai2021joint}
Tuan Lai, Heng Ji, ChengXiang Zhai, and Quan~Hung Tran. 2021.
\newblock Joint biomedical entity and relation extraction with knowledge-enhanced collective inference.
\newblock In \emph{Proceedings of the 59th Annual Meeting of the Association for Computational Linguistics and the 11th International Joint Conference on Natural Language Processing (Volume 1: Long Papers)}, pages 6248--6260.

\bibitem[{Lee et~al.(2020)Lee, Jiang, and Yu}]{lee2020harmonized}
Dongha Lee, Xiaoqian Jiang, and Hwanjo Yu. 2020.
\newblock Harmonized representation learning on dynamic ehr graphs.
\newblock \emph{Journal of biomedical informatics}, 106:103426.

\bibitem[{Li et~al.(2023{\natexlab{a}})Li, Fang, Yang, Wang, Ye, Zhao, and Zhang}]{li2023evaluating}
Bo~Li, Gexiang Fang, Yang Yang, Quansen Wang, Wei Ye, Wen Zhao, and Shikun Zhang. 2023{\natexlab{a}}.
\newblock Evaluating chatgpt's information extraction capabilities: An assessment of performance, explainability, calibration, and faithfulness.
\newblock \emph{arXiv preprint arXiv:2304.11633}.

\bibitem[{Li et~al.(2023{\natexlab{b}})Li, Wang, Zhang, and Zhang}]{li-etal-2023-rethinking}
Jing Li, Yequan Wang, Shuai Zhang, and Min Zhang. 2023{\natexlab{b}}.
\newblock \href {https://doi.org/10.18653/v1/2023.findings-acl.353} {Rethinking document-level relation extraction: A reality check}.
\newblock In \emph{Findings of the Association for Computational Linguistics: ACL 2023}, pages 5715--5730, Toronto, Canada. Association for Computational Linguistics.

\bibitem[{Li et~al.(2021)Li, Li, Wang, Huang, Cho, Ji, Han, and Voss}]{li2021future}
Manling Li, Sha Li, Zhenhailong Wang, Lifu Huang, Kyunghyun Cho, Heng Ji, Jiawei Han, and Clare Voss. 2021.
\newblock The future is not one-dimensional: Complex event schema induction by graph modeling for event prediction.
\newblock In \emph{Proceedings of the 2021 Conference on Empirical Methods in Natural Language Processing}, pages 5203--5215.

\bibitem[{Li et~al.(2022)Li, Wehbe, Ahmad, Wang, and Luo}]{li2022clinical}
Yikuan Li, Ramsey~M Wehbe, Faraz~S Ahmad, Hanyin Wang, and Yuan Luo. 2022.
\newblock Clinical-longformer and clinical-bigbird: Transformers for long clinical sequences.
\newblock \emph{arXiv preprint arXiv:2201.11838}.

\bibitem[{Lim et~al.(2019)Lim, Jeong, and Choi}]{lim2019survey}
Chae-Gyun Lim, Young-Seob Jeong, and Ho-Jin Choi. 2019.
\newblock Survey of temporal information extraction.
\newblock \emph{Journal of Information Processing Systems}, 15(4):931--956.

\bibitem[{Lin et~al.(2016{\natexlab{a}})Lin, Dligach, Miller, Bethard, and Savova}]{lin2016multilayered}
Chen Lin, Dmitriy Dligach, Timothy~A Miller, Steven Bethard, and Guergana~K Savova. 2016{\natexlab{a}}.
\newblock Multilayered temporal modeling for the clinical domain.
\newblock \emph{Journal of the American Medical Informatics Association}, 23(2):387--395.

\bibitem[{Lin et~al.(2016{\natexlab{b}})Lin, Miller, Dligach, Bethard, and Savova}]{lin2016improving}
Chen Lin, Timothy Miller, Dmitriy Dligach, Steven Bethard, and Guergana Savova. 2016{\natexlab{b}}.
\newblock Improving temporal relation extraction with training instance augmentation.
\newblock In \emph{Proceedings of the 15th Workshop on Biomedical Natural Language Processing}, pages 108--113.

\bibitem[{Lin et~al.(2019)Lin, Miller, Dligach, Bethard, and Savova}]{lin2019bert}
Chen Lin, Timothy Miller, Dmitriy Dligach, Steven Bethard, and Guergana Savova. 2019.
\newblock A bert-based universal model for both within-and cross-sentence clinical temporal relation extraction.
\newblock In \emph{Proceedings of the 2nd Clinical Natural Language Processing Workshop}, pages 65--71.

\bibitem[{Lin et~al.(2021)Lin, Miller, Dligach, Bethard, and Savova}]{linentitybert}
Chen Lin, Timothy Miller, Dmitriy Dligach, Steven Bethard, and Guergana Savova. 2021.
\newblock \href {https://doi.org/10.18653/v1/2021.bionlp-1.21} {{E}ntity{BERT}: Entity-centric masking strategy for model pretraining for the clinical domain}.
\newblock In \emph{Proceedings of the 20th Workshop on Biomedical Language Processing}, pages 191--201, Online. Association for Computational Linguistics.

\bibitem[{Liu et~al.(2021)Liu, Xu, Chen, and Zhang}]{liu2021discourse}
Jian Liu, Jinan Xu, Yufeng Chen, and Yujie Zhang. 2021.
\newblock Discourse-level event temporal ordering with uncertainty-guided graph completion.
\newblock In \emph{IJCAI}, pages 3871--3877.

\bibitem[{Liu et~al.(2019)Liu, Wang, Chaudhary, and Liu}]{liu-etal-2019-attention}
Sijia Liu, Liwei Wang, Vipin Chaudhary, and Hongfang Liu. 2019.
\newblock \href {https://doi.org/10.18653/v1/W19-1917} {Attention neural model for temporal relation extraction}.
\newblock In \emph{Proceedings of the 2nd Clinical Natural Language Processing Workshop}, pages 134--139, Minneapolis, Minnesota, USA. Association for Computational Linguistics.

\bibitem[{Liu et~al.(2017)Liu, Yang, Wang, Chen, Tang, Wang, and Xu}]{liu2017entity}
Zengjian Liu, Ming Yang, Xiaolong Wang, Qingcai Chen, Buzhou Tang, Zhe Wang, and Hua Xu. 2017.
\newblock Entity recognition from clinical texts via recurrent neural network.
\newblock \emph{BMC medical informatics and decision making}, 17:53--61.

\bibitem[{Magnini et~al.(2022)Magnini, Altuna, Lavelli, Minard, Speranza, and Zanoli}]{magnini2022european}
Bernardo Magnini, Bego{\~n}a Altuna, Alberto Lavelli, Anne-Lyse Minard, Manuela Speranza, and Roberto Zanoli. 2022.
\newblock European clinical case corpus.
\newblock In \emph{European Language Grid: A Language Technology Platform for Multilingual Europe}, pages 283--288. Springer International Publishing Cham.

\bibitem[{Maldonado et~al.(2019)Maldonado, Yetisgen, and Harabagiu}]{maldonado2019adversarial}
Ramon Maldonado, Meliha Yetisgen, and Sanda~M Harabagiu. 2019.
\newblock Adversarial learning of knowledge embeddings for the unified medical language system.
\newblock \emph{AMIA Summits on Translational Science Proceedings}, 2019:543.

\bibitem[{Mathur et~al.(2021)Mathur, Jain, Dernoncourt, Morariu, Tran, and Manocha}]{mathur2021timers}
Puneet Mathur, Rajiv Jain, Franck Dernoncourt, Vlad Morariu, Quan~Hung Tran, and Dinesh Manocha. 2021.
\newblock Timers: document-level temporal relation extraction.
\newblock In \emph{Proceedings of the 59th Annual Meeting of the Association for Computational Linguistics and the 11th International Joint Conference on Natural Language Processing (Volume 2: Short Papers)}, pages 524--533.

\bibitem[{Mikolov et~al.(2013)Mikolov, Sutskever, Chen, Corrado, and Dean}]{mikolov2013distributed}
Tomas Mikolov, Ilya Sutskever, Kai Chen, Greg~S Corrado, and Jeff Dean. 2013.
\newblock Distributed representations of words and phrases and their compositionality.
\newblock \emph{Advances in neural information processing systems}, 26.

\bibitem[{Miller et~al.(2023)Miller, Bethard, Dligach, and Savova}]{miller2023end}
Timothy Miller, Steven Bethard, Dmitriy Dligach, and Guergana Savova. 2023.
\newblock End-to-end clinical temporal information extraction with multi-head attention.
\newblock In \emph{The 22nd Workshop on Biomedical Natural Language Processing and BioNLP Shared Tasks}, pages 313--319.

\bibitem[{Ning et~al.(2017)Ning, Feng, and Roth}]{ning2017structured}
Qiang Ning, Zhili Feng, and Dan Roth. 2017.
\newblock A structured learning approach to temporal relation extraction.
\newblock In \emph{Proceedings of the 2017 Conference on Empirical Methods in Natural Language Processing}, pages 1027--1037.

\bibitem[{Peng et~al.(2019)Peng, Yan, and Lu}]{peng2019transfer}
Yifan Peng, Shankai Yan, and Zhiyong Lu. 2019.
\newblock Transfer learning in biomedical natural language processing: An evaluation of bert and elmo on ten benchmarking datasets.
\newblock In \emph{Proceedings of the 18th BioNLP Workshop and Shared Task}, pages 58--65.

\bibitem[{Pickhardt et~al.(2021)Pickhardt, Graffy, Perez, Lubner, Elton, and Summers}]{pickhardt2021opportunistic}
Perry~J Pickhardt, Peter~M Graffy, Alberto~A Perez, Meghan~G Lubner, Daniel~C Elton, and Ronald~M Summers. 2021.
\newblock Opportunistic screening at abdominal ct: use of automated body composition biomarkers for added cardiometabolic value.
\newblock \emph{RadioGraphics}, 41(2):524--542.

\bibitem[{Qin et~al.(2023{\natexlab{a}})Qin, Feng, and Van~Durme}]{qin-etal-2023-nlp}
Guanghui Qin, Yukun Feng, and Benjamin Van~Durme. 2023{\natexlab{a}}.
\newblock \href {https://doi.org/10.18653/v1/2023.eacl-main.273} {The {NLP} task effectiveness of long-range transformers}.
\newblock In \emph{Proceedings of the 17th Conference of the European Chapter of the Association for Computational Linguistics}, pages 3774--3790, Dubrovnik, Croatia. Association for Computational Linguistics.

\bibitem[{Qin et~al.(2023{\natexlab{b}})Qin, Feng, and Van~Durme}]{qin2023nlp}
Guanghui Qin, Yukun Feng, and Benjamin Van~Durme. 2023{\natexlab{b}}.
\newblock The nlp task effectiveness of long-range transformers.
\newblock In \emph{Proceedings of the 17th Conference of the European Chapter of the Association for Computational Linguistics}, pages 3774--3790.

\bibitem[{Radford et~al.(2018)Radford, Narasimhan, Salimans, and Sutskever}]{gpt12018}
Alec Radford, Karthik Narasimhan, Tim Salimans, and Ilya Sutskever. 2018.
\newblock \href {https://cdn.openai.com/research-covers/language-unsupervised/language_understanding_paper.pdf} {Improving language understanding by generative pre-training}.

\bibitem[{Rohanian et~al.(2023)Rohanian, Nouriborji, and Clifton}]{rohanian2023exploring}
Omid Rohanian, Mohammadmahdi Nouriborji, and David~A Clifton. 2023.
\newblock Exploring the effectiveness of instruction tuning in biomedical language processing.
\newblock \emph{arXiv preprint arXiv:2401.00579}.

\bibitem[{Saiz and Altuna(2023)}]{saiz2023end}
Jos{\'e}~Javier Saiz and Bego{\~n}a Altuna. 2023.
\newblock End-to-end temporal relation extraction in the clinical domain.
\newblock In \emph{Text2Story@ ECIR}.

\bibitem[{Scheetz et~al.(2021)Scheetz, Koca, McGuinness, Holloway, Tan, Zhu, O’Day, Sandhu, MacIsaac, Gilfillan et~al.}]{scheetz2021real}
Jane Scheetz, Dilara Koca, Myra McGuinness, Edith Holloway, Zachary Tan, Zhuoting Zhu, Rod O’Day, Sukhpal Sandhu, Richard~J MacIsaac, Chris Gilfillan, et~al. 2021.
\newblock Real-world artificial intelligence-based opportunistic screening for diabetic retinopathy in endocrinology and indigenous healthcare settings in {Australia}.
\newblock \emph{Scientific Reports}, 11(1):1--11.

\bibitem[{Shi et~al.(2017)Shi, Li, Zhang, Sun, and Yu}]{hetersurvey2017}
Chuan Shi, Yitong Li, Jiawei Zhang, Yizhou Sun, and Philip~S. Yu. 2017.
\newblock A survey of heterogeneous information network analysis.
\newblock \emph{IEEE Transactions on Knowledge and Data Engineering}, 29(1):17--37.

\bibitem[{Si et~al.(2019)Si, Wang, Xu, and Roberts}]{si2019enhancing}
Yuqi Si, Jingqi Wang, Hua Xu, and Kirk Roberts. 2019.
\newblock Enhancing clinical concept extraction with contextual embeddings.
\newblock \emph{Journal of the American Medical Informatics Association}, 26(11):1297--1304.

\bibitem[{Sohn et~al.(2013)Sohn, Wagholikar, Li, Jonnalagadda, Tao, Komandur~Elayavilli, and Liu}]{sohn2013comprehensive}
Sunghwan Sohn, Kavishwar~B Wagholikar, Dingcheng Li, Siddhartha~R Jonnalagadda, Cui Tao, Ravikumar Komandur~Elayavilli, and Hongfang Liu. 2013.
\newblock Comprehensive temporal information detection from clinical text: medical events, time, and tlink identification.
\newblock \emph{Journal of the American Medical Informatics Association}, 20(5):836--842.

\bibitem[{Styler~IV et~al.(2014)Styler~IV, Bethard, Finan, Palmer, Pradhan, De~Groen, Erickson, Miller, Lin, Savova et~al.}]{styler2014temporal}
William~F Styler~IV, Steven Bethard, Sean Finan, Martha Palmer, Sameer Pradhan, Piet~C De~Groen, Brad Erickson, Timothy Miller, Chen Lin, Guergana Savova, et~al. 2014.
\newblock Temporal annotation in the clinical domain.
\newblock \emph{Transactions of the association for computational linguistics}, 2:143--154.

\bibitem[{Sun et~al.(2013{\natexlab{a}})Sun, Rumshisky, and Uzuner}]{sun2013annotating}
Weiyi Sun, Anna Rumshisky, and Ozlem Uzuner. 2013{\natexlab{a}}.
\newblock Annotating temporal information in clinical narratives.
\newblock \emph{Journal of biomedical informatics}, 46:S5--S12.

\bibitem[{Sun et~al.(2013{\natexlab{b}})Sun, Rumshisky, and Uzuner}]{sun2013evaluating}
Weiyi Sun, Anna Rumshisky, and Ozlem Uzuner. 2013{\natexlab{b}}.
\newblock Evaluating temporal relations in clinical text: 2012 i2b2 challenge.
\newblock \emph{Journal of the American Medical Informatics Association: JAMIA}, 20(5):806.

\bibitem[{Tang et~al.(2013)Tang, Wu, Jiang, Chen, Denny, and Xu}]{tang2013hybrid}
Buzhou Tang, Yonghui Wu, Min Jiang, Yukun Chen, Joshua~C Denny, and Hua Xu. 2013.
\newblock A hybrid system for temporal information extraction from clinical text.
\newblock \emph{Journal of the American Medical Informatics Association}, 20(5):828--835.

\bibitem[{Tourille et~al.(2017)Tourille, Ferret, Neveol, and Tannier}]{tourille2017neural}
Julien Tourille, Olivier Ferret, Aurelie Neveol, and Xavier Tannier. 2017.
\newblock Neural architecture for temporal relation extraction: A bi-lstm approach for detecting narrative containers.
\newblock In \emph{Proceedings of the 55th Annual Meeting of the Association for Computational Linguistics (Volume 2: Short Papers)}, pages 224--230.

\bibitem[{Tvardik et~al.(2018)Tvardik, Kergourlay, Bittar, Segond, Darmoni, and Metzger}]{tvardik2018accuracy}
Nastassia Tvardik, Ivan Kergourlay, Andr{\'e} Bittar, Fr{\'e}d{\'e}rique Segond, Stefan Darmoni, and Marie-H{\'e}l{\`e}ne Metzger. 2018.
\newblock Accuracy of using natural language processing methods for identifying healthcare-associated infections.
\newblock \emph{International Journal of Medical Informatics}, 117:96--102.

\bibitem[{UzZaman and Allen(2011)}]{uzzaman2011temporal}
Naushad UzZaman and James Allen. 2011.
\newblock Temporal evaluation.
\newblock In \emph{Proceedings of the 49th Annual Meeting of the Association for Computational Linguistics: Human Language Technologies}, pages 351--356.

\bibitem[{Vashishtha et~al.(2020)Vashishtha, Poliak, Lal, Van~Durme, and White}]{vashishtha2020temporal}
Siddharth Vashishtha, Adam Poliak, Yash~Kumar Lal, Benjamin Van~Durme, and Aaron~Steven White. 2020.
\newblock Temporal reasoning in natural language inference.
\newblock In \emph{Findings of the Association for Computational Linguistics: EMNLP 2020}, pages 4070--4078.

\bibitem[{Vaswani et~al.(2017)Vaswani, Shazeer, Parmar, Uszkoreit, Jones, Gomez, Kaiser, and Polosukhin}]{transformer2017}
Ashish Vaswani, Noam Shazeer, Niki Parmar, Jakob Uszkoreit, Llion Jones, Aidan~N Gomez, \L~ukasz Kaiser, and Illia Polosukhin. 2017.
\newblock Attention is all you need.
\newblock In \emph{Advances in Neural Information Processing Systems}, volume~30.

\bibitem[{Veličković et~al.(2018)Veličković, Cucurull, Casanova, Romero, Liò, and Bengio}]{gat2018}
Petar Veličković, Guillem Cucurull, Arantxa Casanova, Adriana Romero, Pietro Liò, and Yoshua Bengio. 2018.
\newblock Graph attention networks.
\newblock In \emph{International Conference on Learning Representations}.

\bibitem[{Wang et~al.(2022)Wang, Li, and Xu}]{wang2022dct}
Liang Wang, Peifeng Li, and Sheng Xu. 2022.
\newblock Dct-centered temporal relation extraction.
\newblock In \emph{Proceedings of the 29th international conference on computational linguistics}, pages 2087--2097.

\bibitem[{Wang et~al.(2019)Wang, Ji, Shi, Wang, Ye, Cui, and Yu}]{hgat2019}
Xiao Wang, Houye Ji, Chuan Shi, Bai Wang, Yanfang Ye, Peng Cui, and Philip~S Yu. 2019.
\newblock Heterogeneous graph attention network.
\newblock In \emph{The World Wide Web Conference}, page 2022–2032.

\bibitem[{Xu et~al.(2013)Xu, Wang, Liu, Tsujii, and Chang}]{xu2013end}
Yan Xu, Yining Wang, Tianren Liu, Junichi Tsujii, and Eric I-Chao Chang. 2013.
\newblock An end-to-end system to identify temporal relation in discharge summaries: 2012 i2b2 challenge.
\newblock \emph{Journal of the American Medical Informatics Association}, 20(5):849--858.

\bibitem[{Yan et~al.(2022)Yan, Jia, and Tu}]{yan2022empirical}
Zhaohui Yan, Zixia Jia, and Kewei Tu. 2022.
\newblock An empirical study of pipeline vs. joint approaches to entity and relation extraction.
\newblock In \emph{Proceedings of the 2nd Conference of the Asia-Pacific Chapter of the Association for Computational Linguistics and the 12th International Joint Conference on Natural Language Processing}, pages 437--443.

\bibitem[{Yang et~al.(2020)Yang, Bian, Hogan, and Wu}]{yang2020clinical}
Xi~Yang, Jiang Bian, William~R Hogan, and Yonghui Wu. 2020.
\newblock Clinical concept extraction using transformers.
\newblock \emph{Journal of the American Medical Informatics Association}, 27(12):1935--1942.

\bibitem[{Yao et~al.(2024)Yao, Hochheiser, Yoon, Goldner, and Savova}]{yao-etal-2024-overview}
Jiarui Yao, Harry Hochheiser, WonJin Yoon, Eli Goldner, and Guergana Savova. 2024.
\newblock \href {https://doi.org/10.18653/v1/2024.clinicalnlp-1.53} {Overview of the 2024 shared task on chemotherapy treatment timeline extraction}.
\newblock In \emph{Proceedings of the 6th Clinical Natural Language Processing Workshop}, pages 557--569, Mexico City, Mexico. Association for Computational Linguistics.

\bibitem[{Yu et~al.(2022)Yu, Ji, Li, Ma, Liu, and Xu}]{yu2022s}
Jie Yu, Bin Ji, Shasha Li, Jun Ma, Huijun Liu, and Hao Xu. 2022.
\newblock S-ner: A concise and efficient span-based model for named entity recognition.
\newblock \emph{Sensors}, 22(8):2852.

\bibitem[{Yuan et~al.(2023)Yuan, Xie, and Ananiadou}]{yuan-etal-2023-zero}
Chenhan Yuan, Qianqian Xie, and Sophia Ananiadou. 2023.
\newblock \href {https://doi.org/10.18653/v1/2023.bionlp-1.7} {Zero-shot temporal relation extraction with {C}hat{GPT}}.
\newblock In \emph{The 22nd Workshop on Biomedical Natural Language Processing and BioNLP Shared Tasks}, pages 92--102, Toronto, Canada. Association for Computational Linguistics.

\bibitem[{Zanoli et~al.(2024)Zanoli, Lavelli, do~Amarante, and Toti}]{zanoli2024assessment}
Roberto Zanoli, Alberto Lavelli, Daniel~Verdi do~Amarante, and Daniele Toti. 2024.
\newblock Assessment of the e3c corpus for the recognition of disorders in clinical texts.
\newblock \emph{Natural Language Engineering}, 30(4):851--869.

\bibitem[{Zhang et~al.(2019)Zhang, Song, Huang, Swami, and Chawla}]{hgnn2019}
Chuxu Zhang, Dongjin Song, Chao Huang, Ananthram Swami, and Nitesh~V. Chawla. 2019.
\newblock Heterogeneous graph neural network.
\newblock In \emph{Proceedings of the 25th ACM SIGKDD International Conference on Knowledge Discovery \& Data Mining}, page 793–803.

\bibitem[{Zhang et~al.(2021)Zhang, Huang, and Ning}]{zhang2021extracting}
Shuaicheng Zhang, Lifu Huang, and Qiang Ning. 2021.
\newblock Extracting temporal event relation with syntax-guided graph transformer.
\newblock \emph{arXiv preprint arXiv:2104.09570}.

\bibitem[{Zhang et~al.(2015)Zhang, Hu, Zhang, Mayo, and Chen}]{zhang2015novel}
Yurong Zhang, Gang Hu, Lu~Zhang, Rachel Mayo, and Liwei Chen. 2015.
\newblock A novel testing model for opportunistic screening of pre-diabetes and diabetes among us adults.
\newblock \emph{PLoS One}, 10(3):e0120382.

\bibitem[{Zhong and Chen(2021)}]{zhong2021frustratingly}
Zexuan Zhong and Danqi Chen. 2021.
\newblock A frustratingly easy approach for entity and relation extraction.
\newblock In \emph{Proceedings of the 2021 Conference of the North American Chapter of the Association for Computational Linguistics: Human Language Technologies}, pages 50--61.

\bibitem[{Zhou et~al.(2022)Zhou, Dong, Tu, Wang, and Dou}]{zhou2022rsgt}
Jie Zhou, Shenpo Dong, Hongkui Tu, Xiaodong Wang, and Yong Dou. 2022.
\newblock Rsgt: relational structure guided temporal relation extraction.
\newblock In \emph{Proceedings of the 29th international conference on computational linguistics}, pages 2001--2010.

\bibitem[{Zhou et~al.(2021)Zhou, Yan, Han, Caufield, Chang, Sun, Ping, and Wang}]{zhou2021clinical}
Yichao Zhou, Yu~Yan, Rujun Han, J~Harry Caufield, Kai-Wei Chang, Yizhou Sun, Peipei Ping, and Wei Wang. 2021.
\newblock Clinical temporal relation extraction with probabilistic soft logic regularization and global inference.
\newblock In \emph{Proceedings of the AAAI Conference on Artificial Intelligence}, pages 14647--14655.

\end{thebibliography}

\clearpage
\appendix
\appendix
\setcounter{figure}{0}
\renewcommand{\thefigure}{S\arabic{figure}}

\setcounter{table}{0}
\renewcommand{\thetable}{S\arabic{table}}

\setcounter{section}{0}
\renewcommand{\thesection}{S\arabic{section}}
\setcounter{page}{1}

{\centering
\textbf{\large APPENDIX}\par}
\section{Additional Related Literature}
\label{app:additional_literature}
\paragraph{Temporal Information Extraction.}
 The entity and relation extraction techniques have evolved from machine learning approaches such as Naive Bayes, Markov Logic Network, SVM, CRF, and Structured Perceptron to deep neural networks and recently large pre-trained language models and graph neural networks \citep{lim2019survey, alfattni2020extraction}. The features for training these systems range from rule-based, lexical (uppercase, stopword, sentence lengths, etc.), syntactic and semantic (part-of-speech, tense, dependency paths, etc.), ontology-based (for instance, the top ontology class from Wordnet), to distributional features (word embeddings, pre-trained language model embeddings). For relation extraction, while some works include candidate pair identification and classification both (RE) \citep{xu2013end, tang2013hybrid}, others focus on relation classification (RC), given the candidate pairs \citep{zhou2021clinical}. While some develop separate classifiers for E-E (event-event), E-T (event-timex), and T-T (timex-timex) relations \citep{tang2013hybrid, lin2016multilayered}, others model different subtypes together using the same classifier \citep{zhou2021clinical}. \cite{han2020domain} focus on a sub-problem of end-to-end relation extraction where they only consider E-E (event-event) pairs across short neighbourhoods (3 sentence chunks). Pre-trained word embeddings with additional manually engineered features have been utilized with LSTM \citep{tourille2017neural, liu-etal-2019-attention} and CNN \citep{dligach2017neural} networks on the THYME dataset. 
Subsequently, the introduction of attention mechanism and transformer-based pre-trained language models (PLM) like BERT offered further significant improvements on both entities \citep{alsentzer2019publicly,si2019enhancing,yang2020clinical} and relations \citep{lin2019bert, zhou2021clinical}. The methods for relation classification involve inserting special marker tokens to better represent entity boundaries and processing the sequence using a BERT-based model. This requires creation of such sequences for every entity pair, therefore, at the document level, the task complexity becomes too onerous if the negative samples, i.e. the pairs with No\_Relation are also included. These works therefore remain restricted to the simpler task of relation classification, which excludes the majority No\_Relation class. 
\paragraph{Relation Extraction using Additional Knowledge.} Some of the works also leverage additional knowledge such as temporal calculus \citep{allen1983maintaining} to either infer additional relations or improve consistency of the identified relations \citep{ning2017structured}. \citet{zhou2021clinical} address the relation classification problem and incorporate temporal calculus using probabilistic soft logic (PSL) rules to improve over a simple BERT-based baseline by a significant margin. Others incorporate clinical domain knowledge, such as \citet{xu2013end}, who use additional private and public annotations to improve entity extraction, \citet{lin2016improving}, who enrich the training set using Unified Medical Language System (UMLS) ontology \citep{bodenreider2004unified} and demonstrate improvements on the THYME dataset, others model the data-driven constraints \citep{han2019joint, han2020domain}. More recently, \citet{li-etal-2023-rethinking} leverages the neighboring tokens of each event as local cues and the temporal words between the two events as the global cues to improve the relation classification performance in the general domain. Most of the recent works either perform relation classification given the gold entity pairs or only report performance on a subset of relations such as event-event relations over 1-3 adjacent sentences and ignore the long-distance relations. 
\paragraph{Relation Extraction using LLMs.}
Large language models (LLMs) such as BART \citep{saiz2023end}, T5 \citep{cui2023medtem2}, and Llama2 \citep{rohanian2023exploring} have also been applied to \trex. However, at present, these models perform significantly worse than the large specialized pre-trained model-based approaches within both clinical \citep{saiz2023end} and general domain \trex \citep{gao2023exploring,li2023evaluating}, more so for long-range temporal relations \citep{yuan-etal-2023-zero, chan2023chatgpt}. 
\paragraph{Relation Extraction using \gnns.}
\gnns have been widely applied to heterogeneous graphs, a.k.a knowledge graphs for different applications \citep{hgnn2019, hgat2019, efficientrelation2023}, particularly relation extraction. Prior works use \gnns with document-level graphs using syntactic, semantic, and discourse features for temporal relation classification, given the entities \citep{mathur2021timers, zhou2022rsgt}, or for relation extraction over 1-2 sentences \cite{zhang2021extracting}. \citet{wang2022dct} anchor similarly constructed graphs using the document creation time and obtain considerable improvements. \citet{liu2021discourse} train a \gnn-based model to recover masked edges during training and use the most confident predictions to recover the edges during inference. However, none of these works focus on document-level end-to-end \trex or the clinical domain.

\section{Data}
\label{app:data}
\subsection{I2B2 2012 Corpus}
\paragraph{Pre-processing.}
For the I2B2 2012 corpus, there were several errors and omissions in the gold annotations that were addressed during preprocessing. To begin with, a large number of character offsets required correction when reading from the XML files. Some entity types were not annotated, we omitted these and also the relations involving such omitted entities. Some relations were missing head/tail, these were omitted as well. In several documents, DISCHARGE was not annotated as a SECTIME and still referred to in the TLinks with id ``Discharge''\textemdash this was replaced with the id of an entity with text ``Discharge'' if available, otherwise these relations were omitted as well. 

\begin{figure*}[ht]
    \centering
    \includegraphics[width=.85\linewidth]{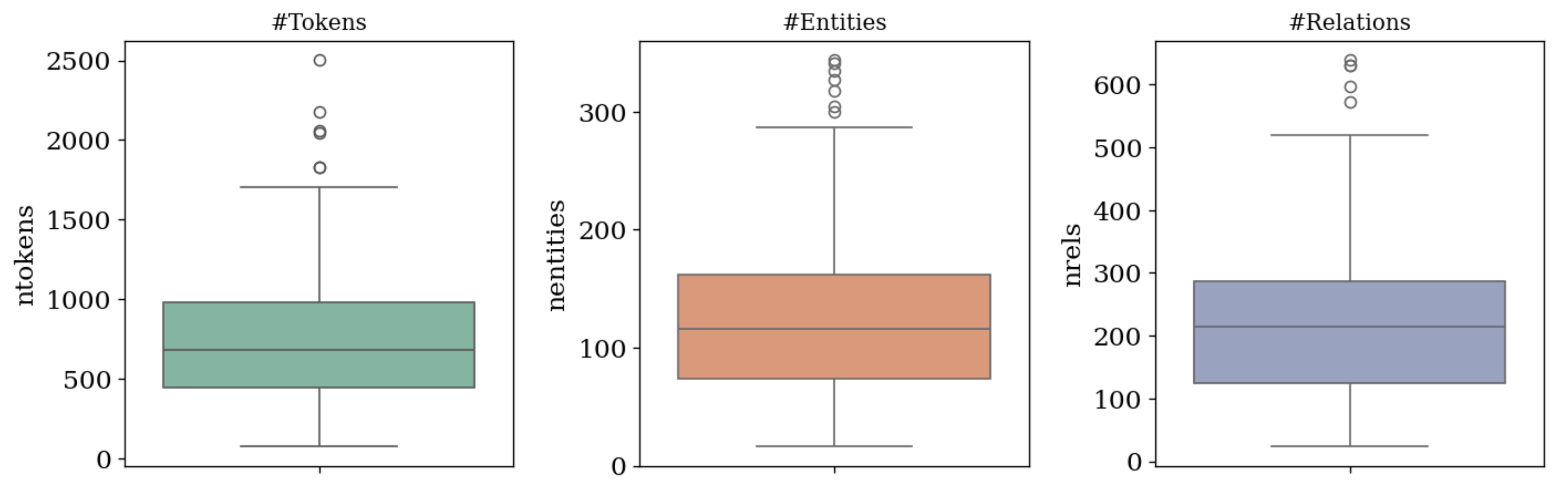}
    \caption{Distributions of the number of tokens, number of entities, and number of relations in each file over the test set in the I2B2 corpus.}
    \label{fig:test_distributions}
\end{figure*}

\paragraph{Additional Data Description.}
We show the distributions of the number of tokens, the number of entities, and the number of relations in each file in the test split in Figure \ref{fig:test_distributions}. On average, each file has 786 tokens, 128 annotated entities, and 229 annotated relation pairs. The maximum token length across the test set is 2503, the maximum number of annotated entities is 344, and the highest number of relations in any file is 639. Note that although relation pairs can be quadratic in terms of number of entities, the annotated relation pairs in the I2B2 dataset are sparse---annotation density 
 is $21\%$ \citep{zhou2021clinical}---due to the difficulty of annotating the long documents. Therefore, to address this, the closure computations are included during the evaluation.

\begin{figure*}[ht]
    \centering
    \includegraphics[width=\linewidth]{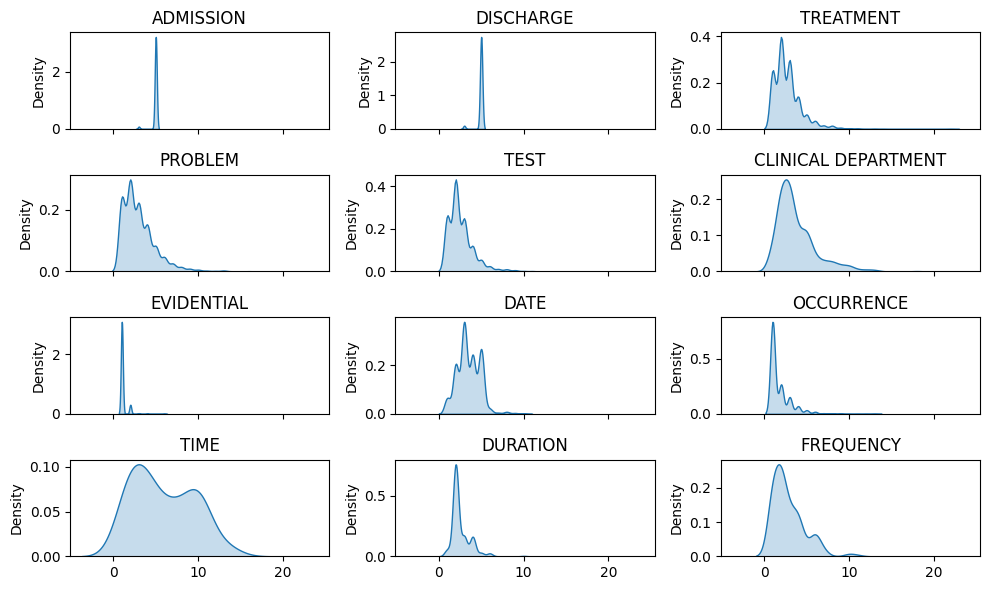}
    \caption{Distributions of the token lengths across each entity type in the test set of the I2B2 corpus.}
    \label{fig:test_entity_length_distribution}
\end{figure*}

\paragraph{Entity Description.} The span-based entity-extraction models start by enumerating all possible contiguous spans in the document. For example, in a document with $n$ tokens $d = [t_1, t_2, t_3, ... t_n]$, the model produces a list of $O(n^2)$ spans. Classifying all possible spans not only increases the computational burden but also increases the data imbalance by introducing many more Not-Entity types, since usually the entity lengths have a much lower bound compared to a document's length. A common way of handling this issue is to place an upper bound on the length of the possible entity spans. We choose a bound of 7 tokens which falls under the 97$^th$ percentile of all entity lengths in the training set. We show the percentile of 7 across each entity type in the training set in Table \ref{tab:etypes}. We also show the distributions of token lengths across each entity type in Figure \ref{fig:test_entity_length_distribution} using kernel density plots as they give a continuous and smooth curve estimating the probability density function. Note that on average, all entity types are shorter than 7 tokens.  The appearance of negative token lengths in the distributions in Figure S2 result of the KDE smoothing process. Only a few instances of type PROBLEM, TEST, TREATMENT, or CLINICAL DEPARTMENT can be longer. For example, the longest token length (using BioMedBERT tokenizer) in the test set for TIME entity is 14 in `5-2-98 , at 4:55 p.m.', for FREQUENCY, it is 11 in `Q4-6H (every 4 to 6 hours )', and for Clinical\_Department, it is 18, and includes a department, physician, and clinic names. We also increased the maximum span length to 20. However, it did not improve the results. 

There are several boundary overlaps among the entities, including overlaps in the same types (e.g., 10/9/2024 vs. 10/9, both annotated as DATE). Therefore, when compiling the gold set, we prioritize the first annotation.  We also experiment with choosing the largest entity span as a tie-breaker strategy for the overlapping spans for training purposes. However, this reduces the performance. On average the entity annotations in the dataset have high label consistency (97.68\% in the training set and 98.13\% in the test set). Entity label consistency is an important measure to validate the agreement in the annotation. Specifically, the span-level entity label consistency  $\phi_{eCon}$ computes whether the same sequence of tokens appearing multiple times in the dataset are annotated with the same entity type, and has a significant impact on an entity classification model's performance \citep{fu2020interpretable}. $\phi_{eCon}(e_{sp})=|y_{sp}==E_{sp}|/|e_{sp}|$, where $E_{sp}$ is the most frequently annotated class for entity span $e_sp$. We report the span-level entity consistency $\phi_{eCon}$ of individual entity classes in Table \ref{tab:etypes}. Note that the label consistency of ADMISSION and DISCHARGE is low and that for DATE is relatively lower than other entity types. Upon further analysis, we find that the ADMISSION/DISCHARGE entities always overlap with the DATE type. Therefore, to address this issue,  we set a priority order to the entity types for label assignment for training purposes. For example, we set the highest priority to ADMISSION/DISCHARGE, then if an entity is annotated with additional labels other than ADMISSION/DISCHARGE, we ignore it. There are numerous other occurrences for DATE type entities and the model can learn to identify DATE types from those. During inference, as an additional post-processing, we add additional DATE type entities for each ADMISSION/DISCHARGE entity.
 \begin{table}[ht]
    \centering
    \adjustbox{max width=.5\textwidth}{
    \begin{tabular}{lccc}
\toprule
 \multirow{2}{*}{\textbf{EType}} &\multirow{2}{*}{Frequency}&Consistency& Percentile\\
 &&$\phi_{eCon}$ (\%)&(7)\\
\midrule
PROBLEM & 4814&99.2&94.8\\
TREATMENT & 3681&98.7&97.2\\
TEST & 2507&98.7&98.2\\
CLINICAL DEPARTMENT & 957&97.9&90.0\\
EVIDENTIAL & 714&90.4&100.0\\
OCCURRENCE & 3122&95.7&99.1\\
ADMISSION & 168&52.3&100.0\\
DISCHARGE & 163&52.6&100.0\\
DATE & 1222&85.0&98.9\\
TIME & 68&97.0&55.9\\
DURATION & 388&95.7&99.1\\
FREQUENCY & 240&98.7&99.6\\
\bottomrule
\end{tabular}
}
\caption{Entity types description for the training set of the I2B2 corpus. Average label consistency over the training set is 97.68, and the percentile of 7 across all entity types is 97.}
    \label{tab:etypes}
\end{table}
\paragraph{Relation Description.} We report the number of relation types across the training, development, and test splits in Table \ref{tab:rtypes}. The data has a high imbalance, with relatively fewer annotations for \textit{After} relations. To address the imbalance during training we flip the relations pairs participating in \textit{Before} to create additional pairs for \textit{After}, and vice-versa. This creates an equal number of \textit{Before}, and \textit{After} relation types for training. However such augmentation introduces some imbalance w.r.t. \textit{Overlap} relation, even after augmenting additional \textit{Overlap} relations by flipping the \textit{Overlap} pairs.
\begin{table}[htbp]
    \centering
    \adjustbox{max width=\textwidth}{
    \begin{tabular}{lccc}
\toprule
 \textbf{Relation Type}&\textbf{Train}&\textbf{Dev} &\textbf{Test}\\
\midrule
\textit{\#Overlap}& 12278 &538& 9889\\
\textit{\#Before}& 16657& 642& 14920\\
\textit{\#After}& 3075&127& 2724\\
   \bottomrule
\end{tabular}
}
  \caption{Number of annotated relations across the training, development, and test set in the I2B2 corpus.}
    \label{tab:rtypes}
\end{table}


\subsection{E3C Corpus}
\label{app:E3C}
\begin{table}[ht]
    \centering
    \adjustbox{max width=\textwidth}{
    \begin{tabular}{lccc}
\toprule
 \textbf{Category}&\textbf{Train}&\textbf{Dev} &\textbf{Test}\\
\midrule
\multicolumn{4}{l}{\textbf{Document Statistics}}\\
\textit{Total Documents}&29&7&48\\
\textit{Average Tokens}&406.14&350.43&392.67\\
\textit{Max Tokens}&797& 694& 828\\
\textit{Min Tokens}& 118& 129& 122\\
\midrule
\multicolumn{4}{l}{\textbf{Event Statistics}}\\
\textit{Total events}&1657&385&2843\\
\textit{Max Token Length}&1.20&1.22&1.24\\
\midrule
\multicolumn{4}{l}{\textbf{Relation Statistics}} \\ 
\textit{\#Overlap}& 962 &220&1560 \\
\textit{\#Before}& 484&102 & 793\\
\textit{\%E-E pairs}&84.81&80.90&85.63\\
   \bottomrule
\end{tabular}
}
  \caption{Descriptive Statistics for the E3C Corpus.}
    \label{tab:e3cstats}
\end{table}
We download the corpus version 2.0.0 from \url{https://live.european-language-grid.eu/catalogue/corpus/7618/download/}. We only use the English subset and the first layer comprising manual annotations. We then select a 20\% random sample of documents from the training set to form the development set and save the best-performing model on this set. 
\paragraph{Pre-processing.} We pre-process the data to adjust the character offsets (primarily due to carriage returns), aligning the text spans with annotated entity names. 

\paragraph{E3C Entity Description.} The dataset has following entity types {BODYPART, PATIENT, EVENT, DATE, OTHER} and timexes including {PREPOSTEXP, DURATION, H-PROFESSIONAL, QUANTIFIER, SET}. However, there are only a few TLinks involving at least one non-EVENT entity (14.37\% in the test set). In contrast, 85.63\% of TLinks are EVENT-EVENT links. Therefore, we only consider EVENT-EVENT TLinks in this work. On average, each EVENT in the training split comprises 1.2 tokens. However, the max token length can be 10. We use the same max span length as I2B2 ($7$) in our experiments which falls under $99.94^{th}$ percentile of training data distribution. 
\paragraph{E3C Relation Description.} The dataset contains both doctime relations and pairwise relations between entities. We only focus on the latter in this work. This subset includes the following relation types: \textit{Before, Contains, Overlap, Begins-On, Ends-On, Simultaneous}. Of these Before and Contains are the most frequent, similar to the I2B2 dataset, we merge the less frequent classes \textit{Overlap, Simultaneous}, with \textit{Contains}, and for clarity, denote it as the Overlap class. We also merge \textit{Ends-On} with \textit{Before}, and \textit{Begins-On} with \textit{Before} after flipping the entity order. 

Descriptive statistics of this dataset are reported in Table \ref{tab:e3cstats}.

\section{Sliding Window Example}
\label{app:slidingwin}
Given a window size of n (e.g. in BERT-based encoders, this can be at most 512), we use a stride of n-2//2 to enumerate overlapping windows. For example, consider the example in Table \ref{tab:slidingwindows}. Given  a sequence with 8 tokens $T_1, T_2, ...T_8$, and let the maximum window size be 6. Each window begins with a CLS and ends with a SEP token. Each token in the window attends to itself and three neighbors. We fix the value of $stride = 6-2//2 =2$, i.e. the window moves 2 tokens at a time. While some tokens participate in overlapping windows, their final embedding is computed from a single window. In the other window, they only provide additional context for other tokens.

We clarify this process with reference to the example in Table \ref{tab:slidingwindows}. The goal is to ensure that each token receives a mask value of $1$ exactly once across all windows. Here mask value 1 means the token contributes to embeddings, while $-2$ means the token embedding is used for context but it doesn't contribute to this token's final representation. Masking is performed in two steps:
\begin{enumerate}
    \item Inner Window Construction:
    \begin{itemize}
        \item Each window has an $inner-window-size = min(\text{window-size}-2, \text{remaining-sequence-length})$. The last window may be shorter and is padded using [PAD] tokens to maintain a consistent size. The stride is set to inner-window-size//2.
    
        \item Initial Mask Assignment: The first window assigns a Mask value of 1 to the leftmost eligible position, while subsequent windows initially assign 0s. The final window assigns a 1 to the rightmost eligible position, while others maintain 0s.
        \item Each window is then divided into three sections:
        \begin{enumerate}
            \item Left section: $inner-window-size//4$ tokens, filled with the leftmost mask value (typically $0$).
            \item Middle section: $inner-window-size//2$ tokens, filled with $1$s (these are the active tokens contributing to final embeddings).
            \item Right section: Remaining positions, filled with the rightmost mask value (typically $0$).
        \end{enumerate}
    \end{itemize}
    \item Prefix and Suffix Application: Special tokens [CLS], [SEP] are respectively added at the start and end of each window with their mask values: $-3$. Any remaining positions are filled with [PAD] token (mask value $= -4$), which is ignored.
\end{enumerate}

This process ensures that in the first window, tokens $T_1, T_2, T_3$ are actively covered ($T_4$ used for context but does not contribute to its own final representation). In the second window, $T_4$ and $T_5$ are covered, followed by $T_6$ and $T_7$ in the third, and in the fourth only $T_8$ is covered. Therefore, each token’s final representation is obtained from a unique window where it actively contributes.

\begin{table}[ht]
    \centering
    \adjustbox{max width=.5\textwidth}{
    \begin{tabular}{cc}
\toprule
\textbf{Token Windows} &{Window Masks}\\
\midrule    
$[CLS, \mathbf{T_1, T_2, T_3}, T_4, SEP]$&$[-3, \mathbf{1, 1, 1}, -2, -3]$ \\
$[CLS, T_3, \mathbf{T_4, T_5}, T_6, SEP]$& $[-3, -2, \mathbf{1, 1}, -2, -3]$\\
$[CLS, T_5, \mathbf{T_6, T_7}, T_8, SEP]$& $[-3, -2, \mathbf{1, 1}, -2, -3]$\\
$[CLS, T_7, \mathbf{T_8}, SEP, PAD, PAD]$ & $[-3, -2, \mathbf{1}, -3, -4, -4]$\\
\bottomrule
\end{tabular}
}
\caption{Illustration of sliding window construction for a sequence $T_1,T_2,... T_8$, with window size $6$. Embeddings of active tokens (in boldface) are used while the others only provide additional context. In addition, the embeddings of CLS tokens in each window are also used to initialize the window nodes.}
\label{tab:slidingwindows}
\end{table}

\section{Graph Construction}
\label{app:graph_details}
We use \spanmodel to construct our initial graph. We begin with mapping each predicted entity type to a node type in the graph. Then we filter all the relations for which the prediction probability $f(\boldsymbol{\epsilon}_{i,j})$ of the relation decoder ($f$ in Eq. \ref{eq:f_eq}) is above a pre-defined threshold $\tau$. We model these high-confidence predictions as edges between the corresponding nodes. This thresholding is of significant importance because we train the \gnn model end-to-end, and our goal is to avoid the propagation of noisy and spurious information through the graph. 

Formally, given the predictions of $g$, we first construct the set of entity nodes $\nodes_{sp} = \{v_i^s \mid i\in [1\cdots N_v], \; \hat{y}^n_{i} \neq Y^n_{none}\}$, where  $Y^n_{none}$:= NOT-ENTITY, and its corresponding set of node types $\ntypes_{sp} = \{\hat{y}^n_i \mid v_i^s\in \nodes_{sp}\}$. We also construct the set of edge types as $\etypes_{sp} = \{(\hat{y}^n_i, \hat{y}^r_{ij}, \hat{y}^n_j) \mid i,j \in [1\cdots N_v],\; i \neq j, \;\hat{y}^r_{ij} \neq Y^r_{none}, \;\hat{y}^n_i, \hat{y}^n_j \in \ntypes_{sp}, \;\max\,\text{softmax}(f(\boldsymbol{\epsilon}_{ij}))\geq \tau\}$, where $Y^r_{none}$:=NO-RELATION. Additionally, we initialize the feature matrix of entity nodes as $X_{sp}=\underset{i=1}{\overset{N_v}{||}}\mathbf{e}_{\text{sp}_{i}}$.

\paragraph{Capturing Context with Global Landmarks. }Additionally, to detect the long-range dependencies, we introduce ``window'' nodes which act as global landmarks. Entity nodes from different windows are connected to their corresponding window node and the window nodes are connected to each other in the lexical order. Formally, for the $k^{th}$ document, we have a set of windows $\nodes_{wn}=\{v^w_l\}_{l=1}^{N_w^k}$, where $N_w^k = \left\lceil \frac{L_k}{L_w} \right\rceil$ shows the number of windows for the $k^{th}$ document of length $L_k$ and $L_w$ is the window size. For this set of windows, we have the set $\ntypes_{wn} = \{\phi(v^w_l) = Y_w^n\}_{l=1}^{N_w^k}$ of window node types, s.t. $Y_w^n$:=WINDOW denotes window node type, and the set of edges to window nodes as $\etypes_{wn} = \{(\hat{y}_i^n, Y_w^e, Y_w^n) \mid \hat{y}^n_i \in \ntypes_{sp}\} \cup \{(\phi(v^w_p), Y_{ww}^e, \phi(v^w_q)) \mid v^w_p, v^w_q \in \nodes_{wn}, \;\text{end}[v^w_p]+1=\text{start}[v^w_q]\}$, where $Y_w^e$:=BELONGS-TO denotes an edge from window node to its descendent entity node, $Y_{ww}^e$:=TO denotes edge type connecting the window nodes, and $\text{start}[.]$ $\text{end}[.]$ denote the start and end indices based on the lexical order of tokens. The feature matrix of window nodes can be written as $X_{wn}=\underset{i=1}{\overset{N_w^k}{||}}\rho_{\text{[CLS]}_i}$, where $\rho_{\text{[CLS]}_i}$ is the BioMedBERT embedding of the [CLS] token for the $i^{th}$ window.


Our experimental results without employing HGT show that capturing the context between two spans probably makes significant improvements because it helps extract the short-range inter-dependencies. Therefore, we also introduce \textit{context} nodes in the graph such that whenever there is a context between two span entities that are not farther than a distance threshold $d_c$, we put a context node between their respective entity nodes. Formally, we add a set of context nodes $\ntypes_{ctx}=\{\phi(v^c_m) = Y_c^n\}_{m=1}^{N_c}$ s.t. $Y_c^n$:=CONTEXT denotes context node type and $N_c$ is the number of context nodes. We make the set of context node edges as $\etypes_{ctx} = \{(\hat{y}_i^n, Y_{cb}^r, Y_c^n), (\hat{y}_j^n, Y_{ca}^r, Y_c^n) \mid i, j\in [1\cdots N_v], \;\hat{y}^n_i,\hat{y}^n_j \in \ntypes_{sp}, \; \max(\text{start}[v^s_i], \text{start}[v^s_j])-\min(\text{end}[v^s_i], \text{end}[v^s_j]) > d_c, \;d_c>1\}$, where $Y_{cb}^r$:=BEFORE-CONTEXT and $Y_{ca}^r$:=AFTER-CONTEXT are before context edge type and after context edge type respectively. For the feature matrix of context nodes we have $X_{ctx}=\underset{i=1}{\overset{N_c}{||}}\mathbf{e}_{\text{ctx}(i,j)}$ s.t. $ i,j\in [1\cdots N_v]$.

At the end, for the final graph, the heterogeneous set of node types is composed as $\ntypes = \ntypes_{sp} \cup \ntypes_{sn} \cup \ntypes_{ctx}$ and the heterogeneous set of edge types is made as $\etypes = \etypes_{sp} \cup \etypes_{wn} \cup \etypes_{ctx}$. Also for the whole feature matrix we have $X=[X_{sp};X_{wn};X_{ctx}]$.

\subsection{HGT Module}\label{app:hgt_module}
We utilize an HGT module to encode the structural information of the document graph in its entities through message-passing (see Figure 2 in \cite{hgt2020} for a representation of this module). This is achieved using transformers \citep{transformer2017} and graph attention networks \citep{gat2018} in three key components: Mutual Attention, Message Passing, and Aggregation. 
\paragraph{Mutual Attention.} Attention scores for aggregating information from node $t$'s neighbors are:
\begin{align}
& \hspace{-2em} Attn(s, e, t) = \underset{s \in \mathcal{N}_t}{\text{softmax}}(\underset{i=1}{\overset{N_h}{||}}A\text{-}head_i(s, e, t))\\ \nonumber
& \hspace{-2em} A\text{-}head_i(s, e, t) = \\ \nonumber
& \frac{\mu_{(\phi(s), \psi(e), \phi(t))}}{\sqrt{d}}(K_i(s) W^{\psi(e)}_a Q_i(t)^T)\\ \nonumber
& \hspace{-2em} K_i(s) = K\text{-}linear_i^{\phi(s)}(h^{(l-1)}(s))\\ \nonumber
& \hspace{-2em} Q_i(t) = Q\text{-}linear_i^{\phi(t)}(h^{(l-1)}(t))
\end{align}
\noindent Where $\mathcal{N}_t$ is the neighborhood of node $t$, $d$ is the embedding size, $A\text{-}head_i(.)$ is the $i^{th}$ attention head, $N_h$ is the number of heads, $W^{\psi(e)}_a \in \realset^{\frac{d}{N_h}\times\frac{d}{N_h}}$ is the edge specific weight matrix and $K\text{-}linear_i^{\phi(s)}: \realset^d \rightarrow \realset^{\frac{d}{N_h}}$ and $Q\text{-}linear_i^{\phi(t)}: \realset^d \rightarrow \realset^{\frac{d}{N_h}}$ denote linear projections specific to the node to compute the key and query. $\mu_{(\phi(s), \psi(e), \phi(t))} \in \realset^{|\ntypes|\times|\etypes|\times|\ntypes|}$ is the significance factor for each meta-relation. These node- and edge-specific projections and attention matrices help with optimal modeling of distribution differences among different types of nodes and edges for information aggregation. 

\paragraph{Message Passing \& Aggregation.}HGT allows edge-type-specific message passing for each node:
\begin{align}
    & \hspace{-2em} Msg(s, e, t) = \underset{i=1}{\overset{N_h}{||}}M\text{-}head_i(s, e, t) \\ \nonumber
    & \hspace{-2em} M\text{-}head_i(s, e, t) = M\text{-}linear_i^{\phi(s)}(h^{(l-1)}(s))W^{\psi(e)}_{m}
\end{align}
\noindent Where, $W^{\psi(e)}_{m}\in \realset^{\frac{d}{N_h}\times\frac{d}{N_h}}$ are edge-specific weights and $M\text{-}linear_i^{\phi(s)}: \realset^d \rightarrow \realset^{\frac{d}{N_h}}$ is node-specific linear projection. Finally, messages are aggregated across all source nodes within the neighborhood of a target node $t$ to update it as:
\begin{align}
& \hspace{-1em} h^{l}(t)=G\text{-}head^{\phi(t)}(\sigma(\tilde{h}^l(t)))+h^{(l-1)}(t) \\
& \hspace{-1em} \tilde{h}^l(t) = \sum_{s\in \mathcal{N}_t}Attn(s, e, t)\cdot Msg(s, e, t)
\end{align}
\noindent Where $\sigma$ is a non-linear function and $G\text{-}head^{\phi(t)}:\realset^d \rightarrow \realset^d$ a linear projection for $t$.

\section{Implementation Details}
\label{app:implementation}

All the experiments have been performed on a single NVIDIA-A100 GPU. For maximum span length, we choose a threshold of 7 tokens that falls within the $97^{th}$ percentile of entity length distribution in the training split. We also experimented with the threshold of 20 which covers 99.99 percentile. Since entities with very long token lengths are rare, we find that 7 gives better performance and is much faster to train. As our base encoder, we use the BioMedBERT variant microsoft/BiomedNLP-BioMedBERT-base-uncased-abstract-fulltext. 

 To obtain span pair representation, we also experiment with concatenating max pooled or mean pooled or attention-based pooling of the embeddings of all the tokens constituting a span. Further, for span pair representation, we experiment with element-wise sum and difference to capture the interaction between entity spans. For incorporating additional context, we also use left-right pooling, where we max-pooled all the tokens within a fixed-sized window to the left and right of each span (with window sizes 10, 100). However, these experiments have lower performance. 

We use the Adam Optimizer with a linear warmup and learning rate of 0.0008 and a dropout of 0.35 (hyperparameter range 0.2-0.35) after the entity and relation classifier. The hyperparameters were manually selected based on some previous works and some experimentation. The weights of the entity and relation decoder are initialized using default uniform distribution. We restrict span-width to a max of 7 with the dimension of the span-width embedding being 7. We also experimented with 1\textendash 2 linear layers with ReLU activation for obtaining span representation and 1\textendash 2 layers for mention and relation decoder. We use 1000 hidden units for each linear layer (we also experimented with 100\textendash 800 units and found the best performance with 1000 units. We are not able to increase these further due to GPU-memory constraints). We use a context window of length $L_w = 512$, this is also the maximum token length the BioMedBERT model can process.

For \gnn-related hyperparameters, we use two attention heads (we experimented with 2\textendash8 heads) and 2 layers (hyperparameter range 1\textendash 3). We run 2 iterations of GNN-based refinement (hyperparameter range = 1\textendash2) and use a dropout of 0.3 after each HGT layer. For selecting the most confident edges from \spanmodel, we use a threshold $\tau$ of 0.4 for the prediction probability (hyperparameter range = 0.1\textendash0.5). For creating dummy nodes, we use a pooling distance $d_c$ = $\delta$*document-length, where $\delta$=0.5 (hyperparameter range 0.5\textendash1). Finally, we use a residual coefficient of 1.0 with the span embeddings to combine them with GNN-refined node embeddings and obtain final entity embeddings (hyperparameter range for residual coefficient = 0.5\textendash 1.0).
 
 We train the model for 20 epochs (we experimented with up to 50 epochs, the models don't show improvement after epoch 20) and save the best-performing model on the validation split after each epoch. We use a batch size of 8 (and experiment with batch sizes in the range 1\textendash16, where 16 is quite large given the GPU memory limitations and quadratic complexity of relation extraction). We allow the base transformer BioMedBERT to fine-tune its weight using a learning rate of 3e-05. We only use the entity loss $\mathcal{L}_{n}$ for the first two epochs and after that use the combined loss $\mathcal{L}= \mathcal{L}_{n} +\mathcal{L}_{r}$ since the model does not generate any entities for the first few epochs. 

 \begin{table}[htbp]
    \centering
    \begin{adjustbox}{width=.5\textwidth}
    \begin{tabular}{lcccc}
    \toprule
         \multirow{2}{*}{Model}&	\multicolumn{2}{c}{Time}&\multirow{2}{*}{\#Parameters}	&Disk Space\\
         \cmidrule(r){2-3}
         &Train (hrs) & Inference (sec)&&(GB)\\
         \midrule 
\spanmodel	&7.3	&14	&1.2M	&0.44\\
\model	&10.8&	14	&1B	&4.35\\
\bottomrule
    \end{tabular}
    \end{adjustbox}
    \caption{Additional Computational Metrics for models trained on the I2B2 corpus}
    \label{tab:appcompute}
\end{table}
 Our method \model takes approximately 11 hours for training and on average 14 seconds for inference on each document, and around 30 minutes on the full test split comprising 120 documents. The computational metrics are summarized in Table \ref{tab:appcompute}.

 For the multi-head attention baseline, we use 128 heads of 64 dimensions and find that the performance decreases by increasing/decreasing the heads.
 
\section{Additional Robustness Checks}
\label{app:additional_exp}
\subsection{Performance Across Additional Encoders and BioMedBERT-UMLS}
\label{app:additionalencoders}
Table \ref{tab:clinical_encoders} reports performance across various encoders pre-trained on clinical texts on the I2B2 corpus. These include MimicEntityBERT \cite{linentitybert}, a fine-tuned variant of BioMedBERT, that uses an entity-masking strategy to mask events and time expressions and achieves SOTA results on the THYME corpus. However, neither this nor other tested models outperform BioMedBERT on NER for the I2B2 2012 dataset. 

We also introduce BioMedBERT-UMLS, a span-based NER model with external knowledge infusion. For this, we integrate concepts extracted from UMLS \citep{bodenreider2004unified} using MetaMap \citep{aronson1994exploiting}. We then construct an induced subgraph of these concepts based on the UMLS semantic network, following \citet{lai2021joint}. Nodes (concepts) are initialized using pre-trained knowledge-infused embeddings from \citet{maldonado2019adversarial}. We model this graph using a Relational Graph Convolutional Network (RGCN) with two-hop message passing. The final node embeddings are concatenated with BioMedBERT embeddings based on span overlap and classified using a two-layer fully connected decoder. This approach sets a new state-of-the-art in timex extraction while achieving comparable performance on event extraction.
\begin{table}[ht]
    \centering
    \adjustbox{max width=.5\textwidth}{
    \begin{tabular}{lcc}
\toprule
Model&	Event&	TimEx \\
&  (F1) &(F1)\\
\midrule
BioClinicalBERT \cite{alsentzer2019publicly}&	86.15	&88.62\\
BlueBERT \cite{peng2019transfer}	&86.51	&88.63\\
EntityBERT \cite{linentitybert}	& 90.18	& 89.50\\
BioMedBERT	&\textbf{90.57}&	90.62\\
BioMedBERT (UMLS-fusion)&	90.05	&\textbf{91.63}\\
\bottomrule
\end{tabular}
}
    \caption{Comparing encoders pre-trained on clinical text on the NER task of the I2B2 dataset with \spanmodel.}
    \label{tab:clinical_encoders}
\end{table}

\begin{table*}[htbp]
    \centering
        \adjustbox{max width=.7\textwidth}{
    \begin{tabular}{l c c cccc c}
\toprule
 \multirow{3}{*}{\textbf{Model}} 
 & \multicolumn{2}{c}{\textbf{EVENT} }
 &\multicolumn{2}{c}{\textbf{TimEx}}
 &\multicolumn{3}{c}{\textbf{TLink}}\\
 \cmidrule(r){2-8}
 & \textbf{EI}&\textbf{EC} & \textbf{EI}&\textbf{EC}&\multicolumn{3}{c}{\textbf{RE}}\\
  \cmidrule{2-8}&$F_{1}$&Acc&$F_{1}$&Acc&P&R&F$_{1}$\\
\midrule              
  \model  & \textbf{89.55}& \textbf{80.99}
                & 90.06& 81.21
                &78.21 & \textbf{61.42} & \textbf{68.81}\\

\midrule
BERT-\spanmodel&	85.87	&71.10	&85.79	&70.93	&75.94&	48.63	&59.30\\
BERT-\model&	85.83	&71.86	&86.25&	72.64	&77.45	&52.30	&62.44\\
\midrule
RoBERTA-\spanmodel&	88.77	&80.34&	\textbf{90.81}&	\textbf{82.58}&	79.98&	52.05&	63.06\\
RoBERTa-\model&	86.42&	75.30&	89.62&	80.05&	78.22&	54.88&	64.50\\
\midrule
Clinical-Longformer&86.62&77.24&91.03&81.76&\textbf{82.39}&52.10&63.84\\
\bottomrule
\end{tabular}
}
    \caption{Expanded Table for Robustness checks with additional encoders for the I2B2 corpus.}
    \label{tab:additional_encoders_full}

\end{table*}
We also present more details of additional encoders presented in Table \ref{tab:additional_encoders} in the expanded Table \ref{tab:additional_encoders_full}, allowing comparisons across span identification (EI) and entity typing (EC).

\subsection{Pipeline Approach.} 
\label{app:pipeline}
We experiment with a pipeline approach that first extracts entities using BioMedBERT-UMLS, a state-of-the-art model for timex extraction, described in the last paragraph. These extracted entities are then passed to an independent relation extraction module, where we apply the method of \citet{zhong2021frustratingly}, which achieves state-of-the-art performance on I2B2 2012 relation classification \citep{cheng2023typed}. 

Unlike end-to-end relation extraction, relation classification is simpler as it only classifies entity pairs that are known to participate in a relation and, therefore, excludes the majority No-relation class representing entity pairs without a clear temporal link. For relation extraction, we add this class by creating multiple negative training samples for each entity pair. Each training sample is constructed by inserting special marker tokens around entities for better-aligning entity boundary representations. While this approach is effective for sentence-level relation extraction, it becomes computationally prohibitive at the document level as it requires generating marker-augmented sentence pairs for every possible entity pair. Therefore, we restrict our experiment to same-sentence relation extraction, focusing on same-sentence entity pairs. Our comparison reveals that this model severely under-performs when tasked with the additional complexity of the No-relation class, with an $F_1$ score of 10.79\%. In comparison, the $F_1$ score of \model on sentence-level relations is 59.67\%.

\subsection{Performance on E-E relations.}
\label{app:ablationee}
As noted by \citet{han2020domain}, predicting E-E TLinks or the temporal relations between two events is more challenging than predicting relations when one of the participating pairs is a timex. This is because the event spans do not explicitly specify the starting or the ending time of an event. We hypothesize that the heterogeneous modeling over various event types helps the model learn the domain-specific constraints. We find that \model indeed improves the performance of E-E relations on the I2B2 corpus with the temporal $F_1$ score over the E-E relations being 49.85\%, while for \spanmodel, the score is 46.27\%. Here, we evaluate these two models over all E-E pairs and not just on the nearby pairs as in \citet{han2020domain}. 

\subsection{Impact of Node Types across E3C}
\begin{table}[htbp]
    \centering
     \adjustbox{max width=\textwidth}{
    \begin{tabular}{l c c}
\toprule
 \multirow{2}{*}{\textbf{Model}} 
 & \textbf{EVENT}  &\textbf{TLink}\\
 \cmidrule(r){2-3}
 & $F_{1}$&$F_{1}$\\
 \midrule
 SPERT &	78.85 &	13.63\\
\spanmodel &	81.30 &22.55 \\
\midrule
\multicolumn{3}{l}{\textbf{\model}}\\
Excluding Window Nodes &\textbf{82.1}	&\textbf{23.48}\\
Excluding Context Nodes & 81.73&23.39\\
Excluding Both & 78.38 &	22.73\\
Including Both Nodes & 81.46 &	22.97\\
\bottomrule
\end{tabular}
}
\caption{The effectiveness of Window and Context nodes on the E3C corpus.}
    \label{tab:e3cablation}
\end{table}
\label{app:ablatione3c}
We also assess the impact of different nodes in HGT construction on the performance across the E3C corpus in Table \ref{tab:e3cablation} and find that Window nodes are not helpful for this corpus. This is expected since, on average, a document in E3C spans only a single window (the maximum window length is 512 tokens, and the average document length in the test set is 392 tokens). Even the longest document in the test set spans only two windows. When we exclude the context nodes but retain the window nodes, these nodes still connect the entities, allowing information propagation. Including both leads to a slight drop in comparison, and excluding both leads to a further marginal drop.

\section{Qualitative Analysis}
\label{app:errors}
\begin{figure*}
    \centering
    \begin{minipage}{0.32\textwidth}
        \centering
        \includegraphics[trim=0 0 480 0, clip, width=\linewidth]{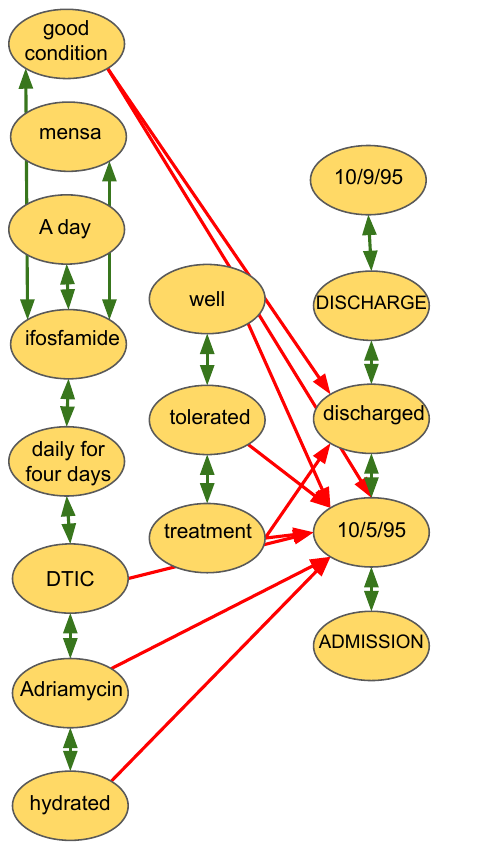}
        \subcaption{GroundTruth}\label{fig:output_sub0}
    \end{minipage}
    \hfill
    \begin{minipage}{0.32\textwidth}
        \centering
        \includegraphics[trim=240 0 240 0, clip, width=\linewidth]{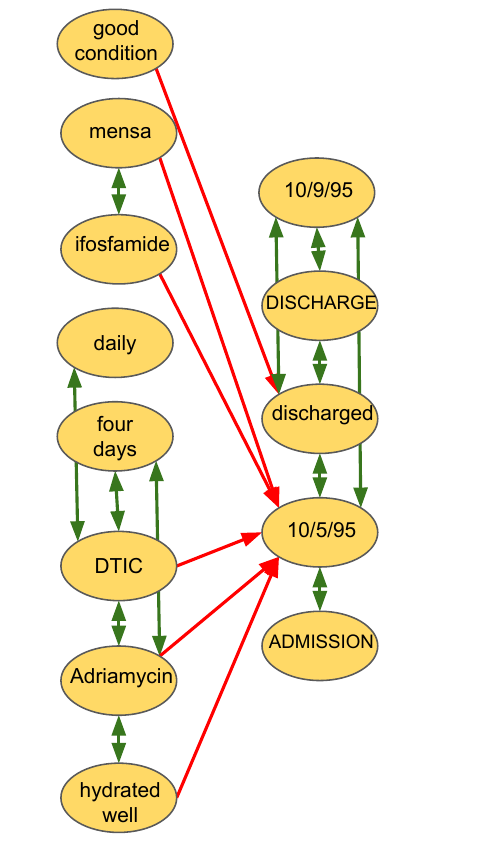}
        \subcaption{\spanmodel}\label{fig:output_sub1}
    \end{minipage}
    \hfill
    \begin{minipage}{0.32\textwidth}
        \centering
        \includegraphics[trim=480 0 0 0, clip, width=\linewidth]{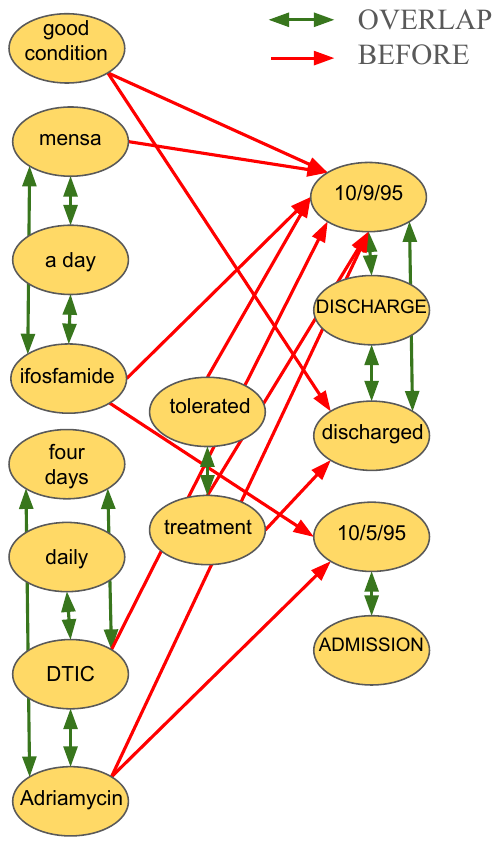}        \subcaption{\model}\label{fig:output_sub2}
    \end{minipage}
    \caption{Ground Truth vs. System Predictions for Temporal Graphs. The figure compares the reference and predicted temporal graphs for a clinical document from the I2B2 corpus.}
    \label{fig:visualization}
\end{figure*}

We perform a qualitative analysis by manually inspecting the output of our \model model. We present an example case study in Figure \ref{fig:visualization} representing one of the documents. To comply with data usage agreements, the original text is not shared, and some node captions are modified while preserving the original meaning. For visualization clarity, we invert the \textit{After} relations to \textit{Before} by flipping the participating entities in the pair. We also remove duplicates after inverting such relations. Note that \model produces more balanced relations compared to \spanmodel which extracts more \textit{Overlap} pairs compared to \textit{Before/After}. Interestingly, there is an error in the ground truth annotations\textemdash the entity `discharged' Overlaps with the admission date. This creates an inconsistency by transitivity, where the dates `10/5/95' and `10/9/95' are overlapping. This issue is also present in \spanmodel predictions. However, \model addresses the issue. For this example, all the graphs demonstrate structural consistency (no cycles involving directed arrows). \spanmodel misses two important wellness indicators\textemdash `tolerated' and `treatment'. Both models also exclude the entity `well', a property of `tolerated' in the text. They also separate `daily for four days' into `daily' and `four days' and connect them with \textit{Overlap} relation, thereby still preserving the meaning. 

We also find that some 'errors' by standard metrics reflect the strengths of our model, revealing gaps in the gold standard. For example, given the co-referring pairs such as (`discharge', `discharged'), (`admission', `hospitalization'), our model can correctly identify the second mention as a relevant entity type even though only the first span has been labeled in the ground truth. This yields additional correct relations with the missing entities that are also missing in the gold standard. Lastly, we note that for some of the documents, the model inconsistently predicts the relation between discharge and admission as \textit{Before}. Approaches to infuse real-world commonsense knowledge to improve this might be an interesting future direction. 

\end{document}